\PassOptionsToPackage{table,xcdraw,usenames,dvipsnames}{xcolor}
\documentclass{article}

\usepackage{microtype}
\usepackage{graphicx}
\usepackage{subfigure}
\usepackage{booktabs} % for professional tables

\usepackage{hyperref}

\usepackage[accepted]{icml2025}

\usepackage{amsmath}
\usepackage{amssymb}
\usepackage{mathtools}
\usepackage{amsthm}

\usepackage[capitalize,noabbrev]{cleveref}

\theoremstyle{plain}
\newtheorem{theorem}{Theorem}[section]

\theoremstyle{definition}
\newtheorem{definition}[theorem]{Definition}

\theoremstyle{remark}

\usepackage[textsize=tiny]{todonotes}

\usepackage{xspace}
\newcommand{\ourmethod}{{{MaAS}}\xspace}
\newcommand{\llmname}[1]{{\fontfamily{pcr}\selectfont {#1}}\xspace}

\usepackage{xcolor}

\definecolor{ForestGreen}{RGB}{34,139,34}
\definecolor{myyellow}{RGB}{181, 181, 27}

\usepackage[table,xcdraw,usenames,dvipsnames]{xcolor}

\newcommand{\blue}[1]{$_{\color{BlueGreen}\downarrow #1}$}
\newcommand{\red}[1]{$_{\color{RedOrange}\uparrow #1}$}

\usepackage{multirow}
\usepackage{makecell}
\usepackage{enumerate}
\usepackage{enumitem}
\usepackage{pifont}

\definecolor{mygrey}{gray}{0.4}

\usepackage[ruled,algo2e,vlined]{algorithm2e}
\renewcommand{\algorithmicrequire}{\textbf{Input:}}  % Use Input in the format of Algorithm
\renewcommand{\algorithmicensure}{\textbf{Output:}} % Use Output in the format of Algorithm
\SetKwInOut{Input}{Input}\SetKwInOut{Output}{Output}
\SetKwComment{Comment}{$\triangleright$\ }{}

\usepackage{listings}

\usepackage[most]{tcolorbox}

\icmltitlerunning{Multi-agent Architecture Search via Agentic Supernet}

\begin{document}

\twocolumn[
\icmltitle{Multi-agent Architecture Search via Agentic Supernet}



\icmlsetsymbol{equal}{*}

\begin{icmlauthorlist}
\icmlauthor{Guibin Zhang}{equal,nus,tongji}
\icmlauthor{Luyang Niu}{equal,tongji}
\icmlauthor{Junfeng Fang}{nus}
\icmlauthor{Kun Wang}{ntu}
\icmlauthor{Lei Bai}{ailab}
\icmlauthor{Xiang Wang}{ustc}
\end{icmlauthorlist}

\icmlaffiliation{nus}{National University of Singapore}
\icmlaffiliation{tongji}{Tongji University}
\icmlaffiliation{ustc}{University of Science and Technology of China}
\icmlaffiliation{ntu}{Nanyang Technological University}
\icmlaffiliation{ailab}{Shanghai AI Laboratory}

\icmlcorrespondingauthor{Kun Wang}{wang.kun@ntu.edu.sg}
\icmlcorrespondingauthor{Lei Bai}{baisanshi@gmail.com}
\icmlkeywords{Machine Learning, ICML}

\vskip 0.3in
]



\printAffiliationsAndNotice{\icmlEqualContribution} % otherwise use the standard text.

\begin{abstract}
Large Language Model (LLM)-empowered multi-agent systems extend the cognitive boundaries of individual agents through disciplined collaboration and interaction, while constructing these systems often requires labor-intensive manual designs. Despite the availability of methods to automate the design of agentic workflows, they typically seek to identify a static, complex, one-size-fits-all system, which, however, fails to dynamically allocate inference resources based on the difficulty and domain of each query. To address this challenge, we shift away from the pursuit of a monolithic agentic system, instead optimizing the \textbf{agentic supernet}, a probabilistic and continuous distribution of agentic architectures. We introduce \textbf{\ourmethod}, an automated framework that samples query-dependent agentic systems from the supernet, delivering high-quality solutions and tailored resource allocation (\textit{e.g.}, LLM calls, tool calls, token cost). Comprehensive evaluation across six benchmarks demonstrates that \ourmethod \textbf{(I)} requires only $6\sim45\%$ of the inference costs of existing handcrafted or automated multi-agent systems, \textbf{(II)} surpasses them by $0.54\%\sim16.89\%$, and \textbf{(III)} enjoys superior cross-dataset and cross-LLM-backbone transferability. The code is available at \url{https://github.com/bingreeky/MaAS}.
\end{abstract}

\vspace{-0.5em}
\section{Introduction}
\label{sec:intro}
\vspace{-0.5em}

Large Language Model (LLM)-based agents \citep{autogpt,babyagi,agentgpt} have made remarkable strides in a spectrum of domains, such as question answering~\citep{zhu2024autotqa}, data analysis~\citep{hong2024datainterpreter,li2024autokaggle}, code generation~\citep{reflexion}, web navigation~\citep{deng2024mind2web}, and data synthesis~\citep{butt2024benchagents}, by equipping LLMs with high-level features, including persona~\citep{multi-persona,chen2024persona}, tools~\citep{shen2024hugginggpt,autogpt}, planning~\citep{ACL2024_AUTOACT-Self-Planning,wu2024Graph4planning,he2023lego}, and memory~\citep{zhong2024memorybank,hatalis2023memorymatter,packer2023memgpt}. Building upon the success of single agents, researchers have demonstrated that combining multiple agents, either cooperatively~\citep{zhuge2024gptswarm} or competitively~\citep{zhao2023competeai}, can surpass the cognitive and intellectual capabilities of individuals~\citep{arXiv2023_MultiAgent-Debate,arXiv2023_MultiAgent-Debate_2,multi-persona,blender,autogen,zhang2024cut}, showcasing the collective intelligence in a society of LLM-agents~\citep{piatti2024societyofagents}.

Early multi-agent systems, such as CAMEL~\citep{NeurIPS2023camel}, AutoGen~\citep{autogen}, and MetaGPT~\citep{meta-gpt}, while delivering specialized capacity, often heavily rely on manual configurations, including prompt engineering, agent profiling, and inter-agent communication pipelines~\citep{qian2024scaling}. This dependency significantly limits the rapid adaptation of multi-agent systems to diverse domains and application scenarios~\citep{tang2023verifai,zhang2024aflow}. More recently, the research community has shifted toward automating multi-agent system design. For instance, DsPy~\citep{khattab2023dspy} and EvoPrompting~\citep{guo2023evoprompt} automate prompt optimization, GPTSwarm~\citep{zhuge2024gptswarm} and G-Designer~\citep{zhang2024gdesigner} optimize inter-agent communication, and EvoAgent~\citep{yuan2024evoagent} and AutoAgents~\citep{chen2023autoagents} self-evolve agent profiling. Nevertheless, they typically focus on automating specific aspects of the system. Subsequently, ADAS~\citep{hu2024adas}, AgentSqure~\citep{shang2024agentsquare}, and AFlow~\citep{zhang2024aflow} broaden the design search space. These state-of-the-art (SOTA) methods optimize a \textit{single, complex (multi-)agent workflow} for a given dataset via different search paradigms, \textit{e.g.}, heuristic search~\citep{hu2024adas}, Monte Carlo tree search~\citep{zhang2024aflow}, and evolution~\citep{shang2024agentsquare}, surpassing the performance of manually designed systems.

\begin{figure*}[t]
\centering
\includegraphics[width=1.0\linewidth]{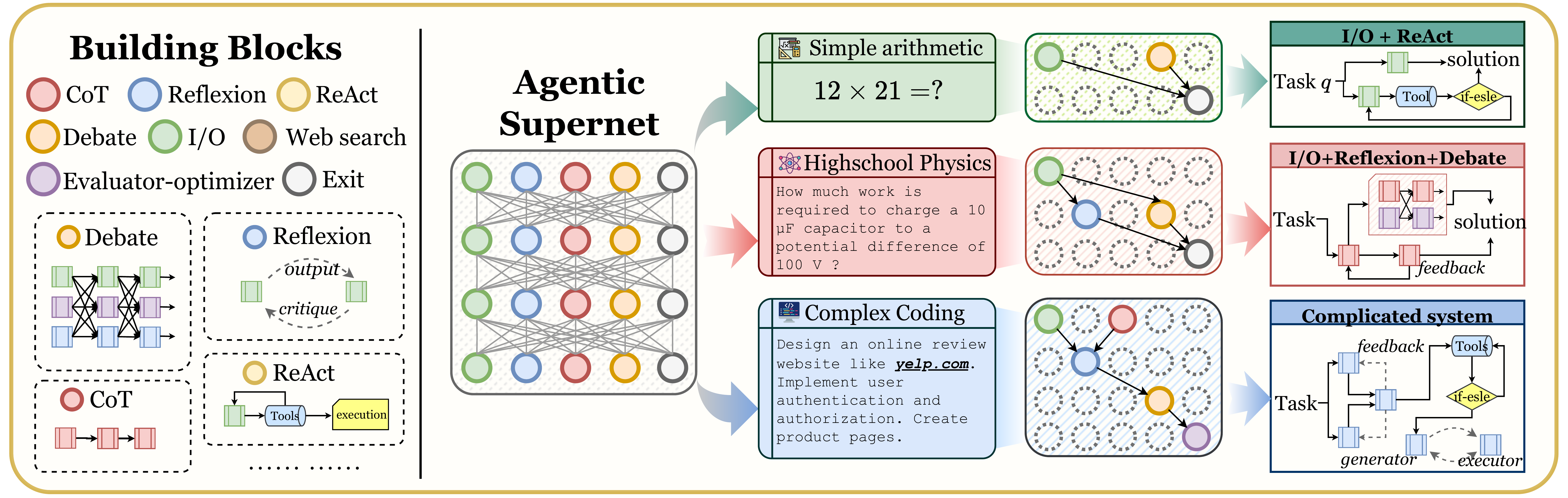}
\vspace{-2em}
\caption{(\textbf{\textit{Left}}) The building blocks of \ourmethod; (\textbf{\textit{Right}}) When confronting different queries, the agentic supernet adaptively samples tailored multi-agent architecture in a query-dependent manner.
}
\vspace{-0.5em}
\label{fig:intro}
\end{figure*}

Although the paradigm of searching for a one-size-fits-all multi-agent system appears sufficient to optimize performance-related metrics such as \textit{accuracy} and \textit{pass@k}, its performance is largely constrained on resource-related metrics, such as token cost, LLM calls, and inference latency (\textbf{Dilemma 1}). Specifically, contemporary methods tend to optimize for a complex and resource-intensive agentic system, often involving dozens of LLM API calls and external tool usage~\citep{arXiv2023_Dynamic-LLM-Agent}. However, this is far from an optimal solution: for example, in mathematical benchmarks~\citep{hendrycksmath2021}, Ph.D.-level abstract algebra may indeed require complicated, token-heavy systems, while simple elementary-level arithmetic works well with a single zero-shot I/O. This paradigm becomes even more problematic when applied to benchmarks across multiple task domains (\textbf{Dilemma 2}): for instance, in the GAIA benchmark~\citep{mialon2023gaiabenchmark}, there is no single system that is optimal for both \textit{file reading} and \textit{web searching} tasks, leaving practitioners with no alternative but to split the benchmark and optimize separately~\citep{zhuge2024gptswarm}. These dilemmas unveil that, the paradigm of automatically optimizing a single multi-agent architecture fails to meet the dynamic and evolving demands of agentic deployment.

To address the above challenges, we propose \textbf{Multi-agent Architecture Search (\ourmethod)}, which, instead of searching for a plausible (possibly non-existent) optimal solution, generates a \textbf{\textit{distribution}} of multi-agent systems. Technically, we model the optimization of \ourmethod on the \textbf{agentic supernet}, a probabilistic, continuous agentic architecture distribution that encompasses a vast number of possible multi-agent candidates. The agentic supernet can be seen as a cascaded multi-layer workflow, including \ding{182} multiple agentic operators (\textit{e.g.}, CoT~\citep{cot},  Multi-agent Debate~\citep{arXiv2023_MultiAgent-Debate}, ReAct~\citep{yao2023react}), as well as \ding{183} the parameterized probability distributions of operators across layers. During training, \ourmethod leverages a \textbf{controller} network to sample multi-agent architectures conditioned on input queries. The distribution parameters and operators are jointly updated based on environmental feedback, with the former's gradients approximated via Monte Carlo sampling and the latter's via textual gradient estimation.
%. The supervision signal from environment feedback can simultaneously update both the distribution parameters and the operators: the former can be easily achieved through gradient backpropagation, while the latter is instantiated by agent-based textual gradients. 
During inference, for different queries, \ourmethod samples a suitable multi-agent system delivering satisfactory resolution and appropriate inference resources, thereby achieving task-customized collective intelligence.

We conduct comprehensive evaluations on seven widely adopted benchmarks, covering diverse use cases in code generation (HumanEval, MBPP), mathematical reasoning (GSM8K, MATH, SVAMP), and diverse tool usage (GAIA). Empirical results demonstrate that \ourmethod is \textbf{\ding{182} high-performing}, surpassing existing handcrafted or automated multi-agent systems by $0.54\%\sim16.89\%$; \textbf{\ding{183} token-economical}, outperforming the SOTA baseline AFlow on the MATH benchmark with $15\%$ of the training cost and $25\%$ of the inference cost; \textbf{\ding{184} transferable} across datasets and LLM-backbones; \textbf{\ding{185} inductive}, demonstrating strong generalizability to unseen agentic operators.

% whose search process is based on the \textbf{agentic supernet}, a probabilistic, continuous agentic architecture distribution that encompasses all possible multi-agent candidates.  在实际部署时，\ourmethod可以基于针对每个user query条件抽样出a suitable multi-agent system with 令人满意的任务解决以及恰到好处的inference resources, 实现task-customized collective intelligence
Briefly put, our key contributions are summarized as follows:
\vspace{-1.8em}
\begin{itemize}[leftmargin=*,itemsep=-0.2em]
\item \textbf{\textit{Paradigm Reformulation}:} We introduce the concept of \textbf{agentic supernet}, a probabilistic, continuous agentic architecture distribution, which transforms the paradigm of optimizing a single optimal multi-agent system into optimizing the distribution of multiple architectures.
\item \textbf{\textit{Practical Solution}:} We propose \ourmethod, an agentic supernet-based framework that automatically evolves powerful multi-agent systems and adaptively allocates high-performing and resource-efficient solutions for user queries with varied difficulty, domain and features.
\item \textbf{\textit{Experimental Evaluation}:} Extensive evaluations on six benchmarks demonstrate that our framework discovers novel agentic systems with $0.54\%\sim16.89\%$ higher performance, significantly lower training/inference costs, transferability across benchmarks and LLMs, and superior inductive capacity.
\end{itemize}
\vspace{-1em}

\begin{figure*}[t]
\centering
\includegraphics[width=1.0\linewidth]{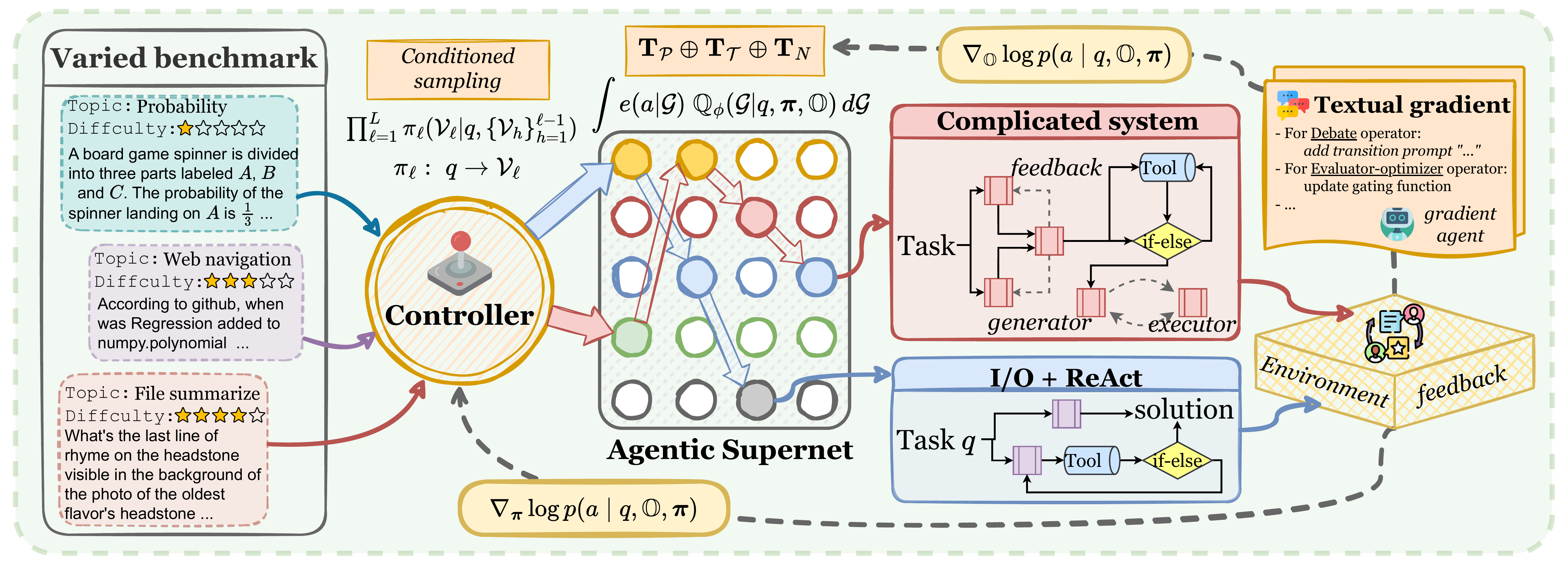}
\vspace{-2em}
\caption{The overall framework of our proposed \ourmethod.
}
\vspace{-0.5em}
\label{fig:framework}
\end{figure*}

% \vspace{-0.5em}
\section{Related Work}
\vspace{-0.3em}
\paragraph{LLM-Agents and Agentic Systems.}
% \vspace{-0.5em}
%With the advent of advanced large language models (LLMs)~\citep{openai2023gpt4,team2023gemini}, significant efforts have been made to develop autonomous agents~\citep{autogpt,babyagi} by equipping LLMs with high-level features, such as persona, tool, planning and memory~\citep{shen2024hugginggpt,zhu2024knowagent,zhong2024memorybank}. 
Building on the success of single agents~\citep{shen2024hugginggpt,zhu2024knowagent,zhong2024memorybank}, studies have shown that grouping multiple LLM-based agents into multi-agent systems (MAS) can substantially enhance individual model capabilities~\citep{FCS2024_Survey-Agent}, as demonstrated in early attempts such as AutoGen~\citep{autogen}, LLM-Debate~\citep{arXiv2023_MultiAgent-Debate}, and AgentVerse~\citep{chen2023agentverse}. %These frameworks significantly expanded the performance boundaries in fields such as code generation~\citep{meta-gpt,chatdev,hu2024evomac}, web navigation~\citep{deng2024mind2web,zhang2024webpilot}, and creative writing~\citep{chen2023autoagents}.
However, they heavily relied on manually crafted designs, which constrained the adaptability and flexibility of agents in addressing unforeseen challenges~\citep{he2023lego,chen2023agentverse}. As a result, automated agentic system design has gained increasing attention in the academic community.

\vspace{-0.6em}
\paragraph{Automating Agentic Systems.}
% \vspace{-0.5em}
Efforts to automate the design of agent-based systems can be broadly classified into the following categories: \textbf{(I) Prompt Optimization}, such as PromptBreeder~\citep{fernando2023promptbreeder}, DsPy~\citep{khattab2023dspy}, and EvoPrompt~\citep{guo2023evoprompt}; \textbf{(II) Inter-agent Communication}, which focuses on orchestrating interactions between agents, including GPTSwarm~\citep{zhuge2024gptswarm}, DyLAN~\citep{arXiv2023_Dynamic-LLM-Agent}, EvoMAC~\citep{hu2024evomac}, AgentPrune~\citep{zhang2024cut} and G-Designer~\citep{zhang2024gdesigner}; and \textbf{(III) Agent Profiling}, represented by AgentVerse~\citep{chen2023agentverse}, EvoAgent~\citep{yuan2024evoagent}, and AutoAgents~\citep{chen2023autoagents}. 
Further, ADAS~\citep{hu2024adas} and AgentSquare~\citep{shang2024agentsquare} provide more comprehensive automation for single-agent design, while AFlow~\citep{zhang2024aflow} achieves multi-agent workflow automation using Monte Carlo tree search (MCTS). However, these high-performing methods still follow the paradigm of searching for a single final system, whereas \ourmethod searches for distribution of architectures with lower average inference costs (LLM calls, token cost, \textit{etc.}).

\vspace{-0.6em}
\paragraph{AutoML.} Automating the design of agentic systems is an emerging topic, yet the history of AutoML~\citep{he2021automl} provides clear precedents. Notably, the progression of agentic automation mirrors that of neural architecture search (NAS)~\citep{ren2021comprehensivenas}. Core NAS techniques, such as reinforcement learning~\citep{zoph2016rl-nas}, evolutionary algorithms~\citep{liu2021evo-nas}, Bayesian optimization (BO)~\citep{white2021bananas}, and MCTS~\citep{wang2021mcts-nas}, have inspired analogous approaches in agentic automation, from policy gradient in \citep{zhuge2024gptswarm} to evolutionary search in \citep{yuan2024evoagent}, BO in \citep{shang2024agentsquare}, and MCTS in \citep{zhang2024aflow}. In NAS, however, these black-box methods were eventually eclipsed by efficient supernet training~\citep{white2023nas1k}, culminating in seminal works like DARTS~\citep{liu2018darts} and SNAS~\citep{xie2018snas}. Inspired by this, we introduce the first MAS searching framework leveraging an \textit{\textbf{agentic supernet}}, posing new paradigms and challenges for agentic automation.

% 在AutoML中，evolutionary algorithm (EA)是被广泛采用的技术手段，在hyperparameter searching以及neural architecture search (NAS)等领域。When it comes to agentic systems，同样存在一些EA-based 早期工作，如

\vspace{-0.4em}
\section{Methodology}
\vspace{-0.4em}
\Cref{fig:framework} illustrates the overall workflow of our method. \ourmethod takes diverse and varying difficulty queries as input and leverages a \textit{controller} to sample a subnetwork from the agentic supernet for each query, corresponding to a customized multi-agent system. After the sampled system executes the query, \ourmethod receives environment feedback and jointly optimizes the supernet’s parameterized distribution and agentic operators. In the following sections, \Cref{sec:prelim} formally defines the search space and optimization objective of \ourmethod, \Cref{sec:arch-sample} details how the controller query-dependently samples multi-agent structures, and \Cref{sec:gradient-update} details the optimization of \ourmethod.

\vspace{-0.4em}
\subsection{Preliminary}\label{sec:prelim}
\vspace{-0.4em}
\paragraph{Search Space.} We first define the basic unit of \ourmethod's search space, namely the agentic operator as follows:
\begin{definition}[\textbf{Agentic Operator}]
\label{def:operator}
% \textit{
An agentic operator $\mathcal{O}$ is a composite LLM-agent invocation process that involves multiple LLM calls and tool usage:
\begin{equation}\label{eq:operator}
\begin{gathered}
\mathcal{O} = \{\{\mathcal{M}_i\}_{i=1}^m, \mathcal{P}, \{\mathcal{T}_i\}_{i=1}^n\},\\
\mathcal{M}_i \in \mathbb{M}, \mathcal{P} \in \mathbb{P}, \mathcal{T}_i \in \mathbb{T},
\end{gathered}
\end{equation}
where $\mathcal{M}$ and $\mathbb{M}$ correspond to LLM backbones and the set of available LLMs, respectively. Similarly, $\mathcal{P}$ and $\mathcal{T}$ represent prompts and tools. $m$ and $n$ denote the number of LLM-agents and tools invoked in the operator, respectively.
% }
\end{definition}
\vspace{-0.3em}
Most existing single/multi-agent workflows can be viewed as agentic operators: CoT~\citep{cot} can be considered one with $m=1$ and $n=0$, denoted as $\mathcal{O}_\text{CoT}$; Self-RAG~\citep{asai2023self-rag} similarly involves $m=1$ agent allocation, but is equipped with $n=1$ retrieval engine, denoted as $\mathcal{O}_\text{SRAG}$; Multi-agent debate~\cite{arXiv2023_MultiAgent-Debate} involves multiple LLM-agent, multi-turn calls, denoted as $\mathcal{O}_\text{Debate}$. The feasible set of agentic operators is denoted as $\mathbb{O}$, and we discuss the initialization of $\mathbb{O}$ in~\Cref{sec:exp-setup,app:operator}. We define a multi-agent system as:
\begin{equation}
\mathcal{G} = \{\mathcal{V},\mathcal{E}\},\;\mathcal{V}\subset\mathbb{O},\;\mathcal{E}\in\mathcal{V}\times \mathcal{V},
\end{equation}
where $\mathcal{V}$ is the set of selected operators in $\mathcal{G}$ and $\mathcal{E}$ denotes their connectivity. $\mathcal{G}$ is constrained as a direct acyclic graph (DAG). Finally, we define the agentic supernet:
% We define the agentic supernet as a probabilistic model representing the hierarchical composition of agentic operators across \(L\) layers. Each layer encodes a probability distribution over all feasible agentic operators, enabling flexible multi-agent system construction. Formally:
\begin{definition}[\textbf{Agentic Supernet}]
\label{def:supernet}
The agentic supernet is denoted as \(\mathcal{A} = \{\boldsymbol{\pi}, \mathbb{O}\}=\{\{\pi_{\ell}(\mathcal{O})\}_{\mathcal{O} \in \mathbb{O}}\}_{\ell=1}^L\), where:
\begin{equation}
\begin{aligned}
\pi_{\ell}(\mathcal{O}) &= p(\mathcal{O} \mid \mathcal{A}_{1:\ell-1}), \;\;\mathcal{O} \in \mathbb{O},\\
\mathcal{A}_{1:\ell-1} &= \{\{\pi_{k}(\mathcal{O})\}_{\mathcal{O} \in \mathbb{O}}\}_{k=1}^{\ell-1},
\end{aligned}
\end{equation}
where \(\pi_{\ell}(\mathcal{O})\) represents the probability of operator \(\mathcal{O}\) present at layer \(\ell\), conditioned on the preceding layers \(\mathcal{A}_{1:\ell-1}\).
The supernet induces a joint distribution over all possible multi-layer operator configurations:
\begin{equation}
p(\mathcal{G}) = \prod_{\ell=1}^L \prod_{\mathcal{O} \in \mathbb{O}} \pi_{\ell}(\mathcal{O})^{\mathbb{I}_{\mathcal{O} \in \mathcal{V}_\ell}},
\end{equation}
where \(\mathbb{I}_{\mathcal{O} \in \mathcal{V}_\ell}\) is the indicator function for the inclusion of \(\mathcal{O}\) in the set of active operators \(\mathcal{V}_\ell\) at layer \(\ell\).
\end{definition}

\vspace{-0.6em}
\paragraph{Problem Formulation.} Given a benchmark $\mathcal{D}$ comprising multiple queries $q$ and their corresponding oracle answers/solutions $a$, the objective of \ourmethod is not to identify a single optimal agentic system like previous practices~\citep{zhang2024aflow,zhuge2024gptswarm}, but to optimize a conditional probability distribution as follows:  
\begin{equation}  
% \small  
 \underset{\mathbb{P}(\mathcal{G} | q)}{\max}\; \mathbb{E}_{\substack{(q,a) \sim \mathcal{D}, \\\mathcal{G} \sim \mathbb{P}(\mathcal{G} | q)}} \big[ U(\mathcal{G}; q,a) - \lambda \cdot C(\mathcal{G}; q) \big],\; \text{s.t.}\;\mathcal{G}\subset\mathcal{A}  
\end{equation}  
where $\mathbb{P}(\mathcal{G} | q)$ is a distribution that generates query-dependent agentic architectures. $U(\cdot)$ and $C(\cdot)$ represent the utiulity/performance and cost of $\mathcal{G}$ for query $q$, respectively, and $\lambda$ is a trade-off parameter.

\vspace{-0.5em}
\subsection{Agentic Architecture Sampling}\label{sec:arch-sample}
\vspace{-0.5em}
The core of \ourmethod lies in tailoring a customized multi-agent system for each user query, which may vary in difficulty and domain, to deliver a satisfactory solution:  
\begin{equation}\label{eq:core}
p(a | q, \boldsymbol{\pi}, \mathbb{O}) = \int e(a | \mathcal{G}) \; \mathbb{Q}_\phi(\mathcal{G} | q, \boldsymbol{\pi}, \mathbb{O}) \, d\mathcal{G},  
\end{equation}  
where $\mathbb{Q}_\phi$ represents the \textit{controller network}, which takes the query $q$, the parameterized distribution $\boldsymbol{\pi}$, and the  available operators $\mathbb{O}$, and outputs the sampled agentic architecture $\mathcal{G}$. $\mathbb{Q}_\phi$ is parameterized by $\phi$, and $e(\cdot|\cdot)$ denotes producing solution via executing $\mathcal{G}$. we implement $\mathbb{Q}_\phi$ as follows:
\begin{equation}
\mathbb{Q}_\phi(\mathcal{G} | q, \boldsymbol{\pi}, \mathbb{O}) = \prod_{\ell=1}^L \pi_\ell(\mathcal{V}_\ell | q, \{\mathcal{V}_h\}_{h=1}^{\ell-1}),
\end{equation}
where $\mathcal{V}_h$ denotes the selected operators at layer $h$. The selection of $\mathcal{V}_\ell$ is conditionally dependent on the query \( q \) and the operators from the previous layers. However, not all queries require execution across \( L \) layers. As discussed in \Cref{sec:intro}, many questions can be resolved with a simple zero-shot I/O~\citep{zhang2024gdesigner}, rendering \( L \) layers unnecessarily redundant. To address this, we introduce an \textbf{early-exit operator}, denoted as \(\mathcal{O}_\text{exit}\). During sampling, if \(\mathcal{O}_\text{exit}\) is encountered, the process exits early:  
\begin{equation}  \label{eq:early-exit}
\begin{aligned}  
\mathbb{Q}_\phi(\mathcal{G} | q, \boldsymbol{\pi}, \mathbb{O}) =& \prod_{\ell=1}^{L} \Big[ \pi_\ell(\mathcal{V}_\ell | q, \{\mathcal{V}_h\}_{h=1}^{\ell-1}) \cdot \mathbb{I}_{\mathcal{O}_{\text{exit}} \notin \mathcal{V}_\ell} \Big] \\ 
&+ \mathbb{I}_{\mathcal{O}_{\text{exit}} \in \mathcal{V}_\ell} \cdot \delta\Big(\ell - \ell_{\text{exit}}\Big),  
\end{aligned}  
\end{equation}  
where \(\ell_\text{exit}\) denotes the layer at which \(\mathcal{O}_{\text{exit}}\) appears, and \(\delta(\cdot)\) is the Kronecker delta function. We implement the sampling process $\pi_\phi$ with a Mixture-of-Expert (MoE)-style network~\citep{shazeer2017outrageously,huang2024harder}:
\begin{equation}\begin{gathered}
\label{eq:layer-l}
\pi_\ell:\;q\rightarrow \mathcal{V}_\ell,\; \mathcal{V}_\ell=\{\mathcal{O}_{\ell 1},\mathcal{O}_{\ell 2},\cdots,\mathcal{O}_{\ell t}\},\\
t = \underset{k\in \{1,\cdots,|\mathbb{O}|\}}{\arg \min} \sum_{j<k}\mathbf{S}^\downarrow_{k}>thres,
\end{gathered}
\end{equation}
where \(\mathbf{S}^\downarrow = \operatorname{sort}(\mathbf{S}, \text{desc})\), and \(\mathbf{S} \in \mathbb{R}^{|\mathbb{O}|} = [S_1, \cdots, S_{|\mathbb{O}|}]\) represents the activation scores of all feasible operators w.r.t. \(q\). Note that \(thres\) is a threshold value that governs operator activation. Operators are activated sequentially, starting from the one with the highest score, and the process continues until the cumulative score exceeds \(thres\). This ensures that the number of selected operators per layer is query-dependent, allowing \ourmethod to dynamically allocate resources based on task complexity.  
\(\mathbf{S}\) is given by: $
S_i = \operatorname{FFN}(\mathbf{v}(q) \| \sum_{\mathcal{O} \in \mathcal{V}_1} \mathbf{v}(\mathcal{O}) \| \cdots \| \sum_{\mathcal{O} \in \mathcal{V}_{\ell - 1}} \mathbf{v}(\mathcal{O})),$
where \(\mathbf{v}(\cdot)\) denotes the embedding function using lightweight models like MiniLM~\citep{wang2020minilm} and SentenceBert~\citep{reimers2019sentence}, and \(\|\) represents concatenation. The detailed implementation of $\mathbf{v}(\cdot)$ is placed in \Cref{app:embedding}.

Upon completing the sequential sampling procedure in \ourmethod, a task-specific multi-agent system \(\mathcal{G}\) is generated and executed to produce the answer \(\widetilde{a}\). In the next section, we elucidate the process of updating the agentic supernet based on environmental feedback.

% 为了做到这一点，对于每个输入的query $q$, \ourmethod leverages a controller network $\mathbb{Q}_\pi(\mathcal{G} \mid q)$:
% \begin{equation}

% \end{equation 

\vspace{-0.5em}
\subsection{Cost-constrained Supernet Optimization}\label{sec:gradient-update}
\vspace{-0.5em}

We present the optimization objective of \ourmethod as follows:
\begin{equation}\label{eq:objective}
\underset{\boldsymbol{\pi},\mathbb{O}}{\min} \;\mathbb{E}_{\substack{(q,a)\sim \mathcal{D}, \mathcal{G}\sim\mathbb{Q}_\phi}} \left[-p(a | q, \boldsymbol{\pi}, \mathbb{O}) + \lambda\cdot C(\mathcal{G};q)\right]
\end{equation}
% \begin{equation}\label{eq:objective}
% \begin{aligned}
% \mathcal{L}(\boldsymbol{\pi}, \mathcal{O}, \mathcal{D}) =& \sum_{\tiny(q,a)\in\mathcal{D}}\!\!\!-\log p(a | q, \boldsymbol{\pi}, \mathbb{O}) \\&+ \lambda\log \mathbb{E}_{\mathcal{G} \sim \mathbb{P}(\mathcal{G} | q)}C(\mathcal{G};q),
% \end{aligned}
% \end{equation}
where \(C(\cdot)\) evaluates the cost of multi-agent systems, represented by token cost, and \(\lambda\) is the trade-off parameter. The term \(p(a | q, \boldsymbol{\pi}, \mathbb{O})\) in \Cref{eq:objective} corresponds to \Cref{eq:core}, where the calculation of \(e(a | \mathcal{G})\) often involves external tools or API-based LLM calls, rendering it non-differentiable. Therefore, we employ an empirical Bayes Monte Carlo procedure~\citep{carlin2000empirical,yan2021fpnas} to estimate the gradient w.r.t the distribution \(\boldsymbol{\pi}\):
\begin{equation}
\label{eq:loss-pi}
\begin{gathered}
\nabla_{\boldsymbol{\pi}}\mathcal{L} \approx \frac{1}{K}\!\!\!\sum_{\scriptscriptstyle(q,a)\in\mathcal{D}}\!\sum_{\scriptscriptstyle k=1}^K\Bigl[m_k\nabla_{\boldsymbol{\pi}}p\left(\mathcal{G}_k\right)\Bigl],\\
m_k=\frac{p(a | q, \mathcal{G}_k)}{\sum_i p(a | q, \mathcal{G}_i)} - \lambda\cdot \frac{C(\mathcal{G}_k;q)}{\sum_i C(\mathcal{G}_i;q)},
\end{gathered}
\end{equation}
where \(m_k\) denotes the cost-aware importance weights of the agentic architecture. Intuitively, the distribution \(\boldsymbol{\pi}\) is updated to favor multi-agent systems that generate high-quality solutions with minimal token cost.

\begin{figure}[!h]
\centering
\vspace{-0.5em}
\includegraphics[width=1.0\linewidth]{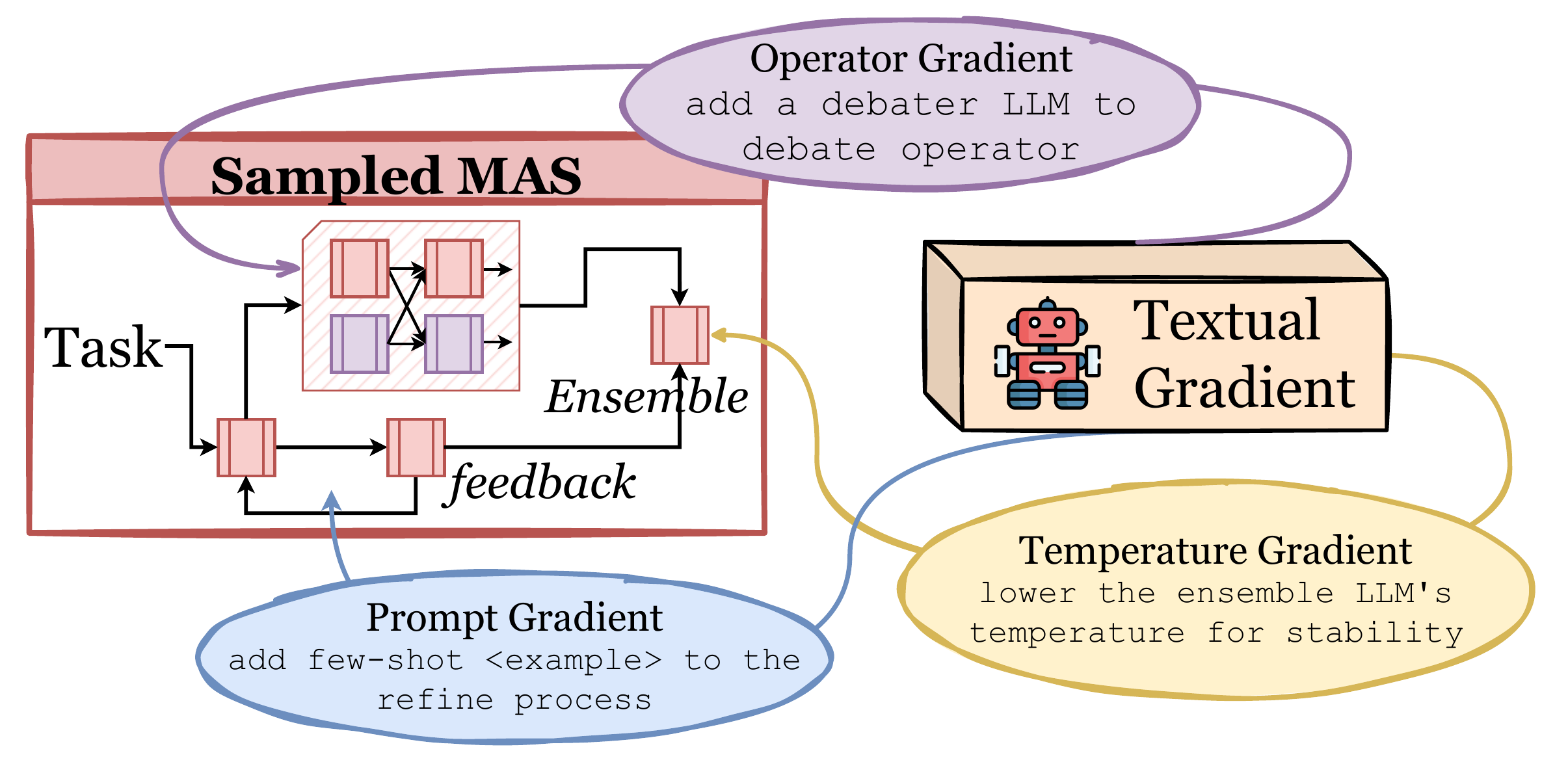}
\vspace{-2em}
\caption{The demonstration of textual gradient.
}
\vspace{-1.em}
\label{fig:gradient}
\end{figure}

However, the gradient w.r.t operators $\nabla_{\mathbb{O}}\mathcal{L}$ cannot be computed similarly. As shown in \Cref{eq:operator}, operators include black-box tool usage and natural language prompts, making numerical gradient updates infeasible. To address this, we utilize agent-based textual gradient ~\citep{chatllm-network,arXiv2023_Dynamic-LLM-Agent,hu2024evomac,zhou2024symbolic-evolve} to approximate the backpropagation for agentic operators, as visualized in \Cref{fig:gradient} and formalized as follows:
\begin{equation}
\label{eq:loss-operator}
\nabla_{\mathbb{O}}\mathcal{L} = \mathbf{T}_{\mathcal{P}} \oplus \mathbf{T}_{\mathcal{T}} \oplus \mathbf{T}_{N}, \mathbf{T}_x \in \mathbb{T}, x\in\{\mathcal{P},\mathcal{T},N\}
\end{equation}
where $\mathbf{T}_{\mathcal{P}}, \mathbf{T}_{\mathcal{T}}, \mathbf{T}_{N}$ represent agent-generated gradient analyses in textual format, corresponding to updates of the prompt, model temperature, and operator node structure (such as merging, splitting, altering, \textit{etc.}), respectively. See prompts in \Cref{app:text-gradient}. In this way, the core components of the agentic supernet, namely the agentic operators and their connectivity, are jointly updated, enabling the fully automated evolution of multi-agent systems. We summarize the notations in \Cref{tab:notations}, and the algorithm in \Cref{alg:algo}.

\begin{algorithm}[!t]
\caption{Algorithm workflow of \ourmethod}\label{alg:algo}
\Input{A dataset $\mathcal{D}$ containing training set $\mathcal{D}_\text{train}$ and test set $\mathcal{D}_\text{test}$, Operator set $\mathbb{O}$, Randomly initialized distribution $\boldsymbol{\pi}$, Controller network $\mathbb{Q}_\phi$}
                
\Output{Well-optimized agentic supernet, composed of distribution $\boldsymbol{\pi}$ and operators $\mathbb{O}$}

% \Output{Optimized operators $\mathbb{O}$ and distribution $\boldsymbol{pi}$}
\For{$(q,a)$ \rm{in} $\mathcal{D}_\text{train}$}{
\tcc{\textcolor{blue}{Sample query-dependent MAS}}
% $\mathcal{G}\leftarrow \mathbb{Q}_\phi(\mathcal{G} | q, \boldsymbol{\pi}, \mathbb{O})$\Comment*[r]{\textcolor{blue}{Eq.~\ref{eq:early-exit}}}
\For{\rm{layer} $\ell \leftarrow 1$ \KwTo $L$}{
$\mathcal{V}_\ell\leftarrow\pi_\phi(\mathcal{V}_\ell | q, \{\mathcal{V}_h\}_{h=1}^{\ell-1})$\Comment*[r]{\textcolor{blue}{Eq.~\ref{eq:layer-l}}}

\If{$\ell = L$ \rm{or} $\mathcal{O}_\text{exit} \in \mathcal{V}_\ell$}{
break\tcp{\textcolor{blue}{Exit when reaching maximal sampling depth or encountering the early-exit operator}}
}
}
Obtain $\mathcal{G}\leftarrow\langle \mathcal{V}_1,\cdots,\mathcal{V}_\ell\rangle$ for query $q$\Comment*[r]{\textcolor{blue}{Eq.~\ref{eq:early-exit}}}
\tcc{\textcolor{blue}{Execute sampled MAS}}
Execute $\mathcal{G}$ and obtain $\Tilde{a}\leftarrow e(a|\mathcal{G})$\Comment*[r]{\textcolor{blue}{Eq.~\ref{eq:core}}}
\tcc{\textcolor{blue}{Self-evolve agentic supernet}}
Compute loss w.r.t. $\boldsymbol{\pi}$, $\nabla_{\boldsymbol{\pi}}\mathcal{L}$\Comment*[r]{\textcolor{blue}{Eq.~\ref{eq:loss-pi}}}

 Estimate loss w.r.t $\mathbb{O}$ via textual gradient\Comment*[r]{\textcolor{blue}{Eq.~\ref{eq:loss-operator}}}

 Update $\boldsymbol{\pi}$ and $\mathbb{O}$ accordingly\Comment*[r]{\textcolor{blue}{Eq.~\ref{eq:objective}}}

}
\end{algorithm}
\vspace{-0.5em}

\begin{table*}[!t]
\centering
\caption{Performance comparison with single agent, hand-craft multi-agent systems, and automated agentic workflows. The base LLM is consistently set as \llmname{gpt-4o-mini} for all baselines. We \textbf{bold} the best results and \underline{underline} the runner-ups.}
\label{tab:rq1_performance}
\renewcommand\tabcolsep{8pt}
\renewcommand\arraystretch{1.1}
  
\resizebox{\linewidth}{!}{
\begin{tabular}{l|cccccc}
\Xhline{1.2pt}
\rowcolor{CadetBlue!20} 
{\textbf{Method}} & \textbf{GSM8K} & \textbf{MATH} & \textbf{MultiArith} & \textbf{HumanEval} & \textbf{MBPP}  & {\textbf{Avg.}} \\
\Xhline{1.2pt}
Vanilla  & $87.45$ & $46.29$ & $96.85$ & $87.08$ & $71.83$ & $77.50$ \\
\hline

\rowcolor{gray!10}CoT~\citep{cot} & $87.10$\blue{0.35} & $46.40$\red{0.11} & $96.31$\blue{0.54} & $88.13$\red{1.05} & $71.83$\blue{0.00} & $77.95$ \\

ComplexCoT~\cite{fu2022complexity}  & $86.89$\blue{0.56} & $46.53$\red{0.24} & $96.70$\blue{0.15} & $87.49$\red{0.41} & $72.36$\red{0.53} & $78.00$ \\

\rowcolor{gray!10}SC (CoT$\times 5$)~\citep{wang2023selfconsistency}  & $87.57$\red{0.12} & $47.91$\red{1.62} & $96.58$\blue{0.27} & $88.60$\red{1.52} & $73.60$\red{1.77} & $78.85$  \\

\hline

MultiPersona~\citep{multi-persona} & $87.50$\red{0.05} & $45.43$\blue{0.86} & $97.49$\red{0.64} & $88.32$\red{1.24} & $73.19$\red{1.36} & $78.39$ \\

\rowcolor{gray!10}LLM-Debate~\citep{arXiv2023_MultiAgent-Debate} & $89.47$\red{2.02} & $48.54$\red{2.25} & $97.33$\red{0.48} & $88.68$\red{1.60} & $70.29$\blue{1.54} & $78.86$ \\

LLM-Blender~\citep{blender} & $88.35$\red{0.90} & $46.92$\red{0.63} & $97.29$\red{0.44} & $88.80$\red{1.72} & $77.05$\red{5.22} & $79.68$ \\

\rowcolor{gray!10}DyLAN~\citep{arXiv2023_Dynamic-LLM-Agent}  & $89.98$\red{2.53} & $48.63$\red{2.34} & $97.12$\red{0.27} & $90.42$\red{3.34} & $77.30$\red{5.47} & $80.69$ \\ 

AgentVerse~\citep{chen2023agentverse}  & $89.91$\red{2.46} & $47.35$\red{1.06} & $97.50$\red{0.65} & $89.29$\red{2.21} & $74.28$\red{2.45} & $79.67$ \\ 

\rowcolor{gray!10}MacNet~\citep{qian2024scaling}  & $87.95$\red{0.50} & $45.18$\blue{1.11} & $96.03$\blue{0.82} & $84.57$\blue{2.51} & $65.28$\blue{6.55} & $75.00$ \\ 

\hline

AutoAgents~\citep{chen2023autoagents}  & $87.69$\red{0.24} & $45.32$\blue{0.97} & $96.42$\blue{0.43} & $87.64$\red{0.56} & $71.95$\red{0.12} & $77.80$ \\

\rowcolor{gray!10}GPTSwarm~\citep{zhuge2024gptswarm}  & $89.14$\red{1.69} & $47.88$\red{1.59} & $96.79$\blue{0.06} & $89.32$\red{2.24} & $77.43$\red{5.60} & $80.11$  \\

ADAS~\cite{hu2024adas}  & $86.12$\blue{1.33} & $43.18$\blue{3.11} & $96.02$\blue{0.83} & $84.19$\blue{2.89} & $68.13$\blue{3.70} & $75.13$  \\

\rowcolor{gray!10}AgentSquare~\cite{shang2024agentsquare} & $87.62$\red{0.17} & $48.51$\red{2.22} & \underline{$97.77$}\red{0.92} & {89.08}\red{2.00} & $78.46$\red{6.63} & {$80.29$} \\

AFlow~\citep{zhang2024aflow}  & \underline{$91.16$}\red{3.71} & \underline{$51.28$}\red{4.91} & $96.22$\blue{0.63} & \underline{$90.93$}\red{3.85} & \underline{$81.67$}\red{9.84} & \underline{$82.25$}  \\

\hline

\rowcolor{gray!10}\textbf{\ourmethod (Ours)}  & {$\mathbf{92.30}$}\red{4.85} & \textbf{$\mathbf{51.82}$}\red{5.53} & \textbf{$\mathbf{98.80}$}\red{1.95} & \textbf{$\mathbf{92.85}$}\red{5.77} & \textbf{$\mathbf{82.17}$}\red{10.34} & \textbf{$\mathbf{83.59}$} \\

\Xhline{1.2pt}
\end{tabular}
}
\vspace{-1.5em}
\end{table*}

\vspace{-1em}
\section{Experiments}
% In this section, we conduct extensive experiments to answer the following research questions ($\mathcal{RQ}$): 
% \vspace{-0.8em}
% \begin{enumerate}[start=1,label={\bfseries $\mathcal{RQ}$\arabic*:},leftmargin=3em,itemsep=-1mm]
% \item Can \ourmethod automatically evolve and design high-performing multi-agent systems?
% \item Does \ourmethod simultaneously maintain low training and inference costs (\textit{e.g.}, token, wall-clock time)?
% \item Is \ourmethod  capable of sampling customized multi-agent systems in a query-aware manner?
% \item How sensitive is \ourmethod to its key hyperparameters and components?
% \end{enumerate}

\vspace{-0.6em}
\subsection{Experiment Setup}\label{sec:exp-setup}
\vspace{-0.4em}
\paragraph{Tasks and Benchmarks.} We evaluate \ourmethod on six public benchmarks covering three domains: \textbf{(1) math reasoning}, GSM8K~\citep{gsm8k}, MATH~\citep{hendrycksmath2021}, and MultiArith~\cite{roy2016solving};  \textbf{(2) code generation}, HumanEval~\citep{human-eval} and MBPP~\citep{austin2021mbpp}); and \textbf{(3) tool use}, GAIA~\citep{mialon2023gaiabenchmark}. For the MATH benchmark, we follow \cite{hong2024datainterpreter} in selecting 617 problems from four typical problem types (Combinatorics \& Probability, Number Theory, Pre-algebra, Pre-calculus). The dataset statistics are in \Cref{app:dataset}.
\vspace{-0.8em}
\paragraph{Baselines.}  We compare \ourmethod with three series of agentic baselines: \textbf{(1) single agent execution methods}, including CoT~\cite{cot}, ComplexCoT~\cite{fu2022complexity}, Self-Consistency~\cite{wang2023selfconsistency}; \textbf{(2) hand-craft multi-agent systems}, including MultiPersona~\citep{multi-persona}, LLM-Debate~\citep{arXiv2023_MultiAgent-Debate}, LLM-Blender~\cite{blender}, DyLAN~\citep{arXiv2023_Dynamic-LLM-Agent}, AgentVerse~\citep{chen2023agentverse} and MacNet~\citep{qian2024scaling}; \textbf{(3) (partially or fully) autonomous multi-agent systems}, including GPTSwarm~\citep{zhuge2024gptswarm}, AutoAgents~\citep{chen2023autoagents}, ADAS~\citep{hu2024adas}, AgentSquare~\citep{shang2024agentsquare} and AFlow~\citep{zhang2024aflow}. More details on baseline setups are provided in \Cref{app:baselines}.

\vspace{-0.8em}
\paragraph{Implementation details.} We leverage both close-source LLM (\llmname{gpt-4o-mini-0718}~\citep{OpenAI-gpt4o}) and open-source LLM (\llmname{Qwen-2.5-72b-instruct}~\cite{yang2024qwen25} and \llmname{llama-3.1-70b}~\cite{dubey2024llama}). All models are accessed via APIs with the temperature set to $1$. We set the number of layers as $L=4$, the cost penalty coefficient $\lambda$ as $\lambda\in\{1e-3,5e-3,1e-2\}$, and the sampling times $K=4$. $thres=0.3$ for \Cref{eq:layer-l}.

\begin{table}[!t]
% \centering
% \label{tab:main_results}
\caption{Performance on GAIA benchmark. {The best and runner-up results are \textbf{bolded} and \underline{underlined}, respectively.}}
% 146 245 75 ｜ 466
% \vspace{-0.5em}
\label{tab:rq1_gaia}
\renewcommand\tabcolsep{8pt}
\renewcommand\arraystretch{1.1}
\centering
  \footnotesize 
\begin{tabular}{l|cccc} 
    \Xhline{1.2pt}
    \rowcolor{CadetBlue!20} 
    \textbf{Method} & \textbf{Level 1} & \textbf{Level 2} & \textbf{Level 3}& \textbf{Avg.} \\
    \Xhline{1pt}
    % GPT-4 \\
    GPT-4o-mini  & $7.53$ & $4.40$ & $0$ & $4.65$\\
   \rowcolor{gray!10} GPT-4 & $9.68$ & $1.89$ & $2.08$ & $4.05$ \\
    \hline
    AutoGPT & $13.21$ & $0$ & $3.85$ & $4.85$\\
    \rowcolor{gray!10}TapeAgent & \underline{$23.66$} & $14.47$ & $\mathbf{10.20}$ & $16.61$ \\
   % \rowcolor{gray!10} DAS Agent\\
   Sibyl & $21.51$ & $15.72$ & $4.08$  & $15.61$\\
    \hline 
    \rowcolor{gray!10}AutoAgents & $16.13$ & $0$ & $0$ & $5.16$  \\
    GPTSwarm & $23.66$ & $16.35$ & $2.04$ &  \underline{$16.33$}\\
    \rowcolor{gray!10}ADAS & $13.98$ & $4.40$ & $0$ & $6.69$ \\
    AgentSquare & $22.58$ & $15.72$ & $6.25$ & \underline{$16.34$} \\
    \rowcolor{gray!10}AFlow & $10.75$ & $8.81$ & $4.08$ & $8.00$  \\
    
    \hline
    \ourmethod  & \textbf{$\mathbf{25.91}$} & \textbf{$\mathbf{22.01}$} & \underline{$6.25$} &  $\mathbf{20.69}$ \\
   % \rowcolor{gray!10}\ourmethod-R  & {\underline{90.30}} & {\color{RedOrange} $\uparrow$\underline{18.62}}\\
    \Xhline{1.2pt}
\end{tabular}

  \vspace{-0.9em}

\vspace{-1.em}
\end{table}

\vspace{-0.4em}
\subsection{Performance Analysis}
\vspace{-0.4em}
We compare \ourmethod with 14 baselines on the GSM8K, MATH, MultiArith, HumanEval, and MBPP benchmarks in \Cref{tab:rq1_performance}, and with 10 baselines on GAIA in \Cref{tab:rq1_gaia}. The following observations can be made:

% \textbf{Obs.\ding{182} Automated agentic systems consistently outperform handcrafted ones.} As shown in \Cref{tab:rq1_performance}, both predefined and automated agentic systems exhibit substantial performance improvements compared to the vanilla LLM backbone (\textit{e.g.}, a $1.60\%\uparrow$ improvement for LLM-Debate and $3.85\%\uparrow$ for AFlow on HumanEval). Furthermore, the improvements of automated systems surpass those of predefined ones. For instance, on GSM8K, the best predefined agentic system, DyLAN, achieves an accuracy of $89.98\%$, while the fully automated AFlow reaches $91.16\%$. 

\begin{table*}[!t]
\centering
\caption{Efficiency comparison between \ourmethod and state-of-the-art baselines on the MATH Benchmark. We shade the values of the lowest token/cost/wall-clock time and the highest performance.}
% \vspace{-0.1em}
\setlength\tabcolsep{3pt}
\resizebox{1\textwidth}{!}{
\begin{tabular}{lccccccccc}
\Xhline{1.2pt} % & \multirow{2}{*}{Methods}
Method & \multicolumn{4}{c}{\textbf{Training}
} & \multicolumn{4}{c}{\textbf{Inference}} & \multicolumn{1}{c}{\textbf{Overall}}    \\
\cmidrule(lr){2-5} \cmidrule(lr){6-9}  \cmidrule(lr){10-10}
  & {\makecell{Prompt\\token}} & {\makecell{Completion\\token}}  & {\makecell{Total\\cost (\$)}} & {\makecell{Wall-clock\\time (min)}}  & {\makecell{Prompt\\token}} & {\makecell{Completion\\token}}  & {\makecell{Total\\cost (\$)}}  & {\makecell{Wall-clock\\time (min)}}  & {\makecell{Acc.\\(\%)}} \\
% \Xhline{1.pt}
\hline 
LLM-Debate & - & - & - & - & $3,275,764$ & $10,459,097$ & $6.76\$ $ & $92$ & $48.54$ \\
DyLAN & $22,152,407$ & $16,147,052$ & $13.01\$$ & $508$ & $6,081,483$ & $3,303,522$ & $2.89\$ $ & $39$ & $48.63$\\
MacNet  & - & - & - & - & $7,522,057$ & $2,043,600$ & $2.35\$ $ & $47$ & $45.18$\\

\hline
GPTSwarm & $21,325,266$ & $6,369,884$ & $7.02\$$ & $129$ & $3,105,571$ & \cellcolor{gray!25}$788,273$ & $0.93\$$ & $30$ & $47.88$\\
AFlow & $33,831,239$ & $29,051,840$ & $22.50\$$ & $184$ & $2,505,944$ & $2,151,931$  & $1.66\$$ & $23$ & $51.28$\\
\hline
\ourmethod & \cellcolor{gray!25}$3,052,159$ & \cellcolor{gray!25}$2,380,505$ & \cellcolor{gray!25}$3.38\$$ & \cellcolor{gray!25}$53$& \cellcolor{gray!25}$1,311,669$ & $853,116$ & \cellcolor{gray!25}$0.42\$$ & \cellcolor{gray!25}$19$  & \cellcolor{gray!25}$51.82$\\

% \midrule
\Xhline{1.2pt}
\end{tabular}
}
\label{tab:efficiency}
\vspace{-0.9em}
\end{table*}

\textbf{Obs.\ding{182} \ourmethod achieves optimal performance across all task domains.} The multi-agent system optimized by \ourmethod outperforms manually designed methods by an average of $3.90\sim6.40\%$ and existing automated methods by $2.07\sim8.26\%$. Overall, as for mathematical reasoning and code generation, \ourmethod achieves an average best score of $83.59\%$, demonstrating its versatility and superiority. \Cref{tab:rq1_gaia} shows a comparison of \ourmethod with automated systems and three additional baselines, including AutoGPT~\citep{autogpt}, TapeAgent~\citep{bahdanau2024tapeagents}, and Sibyl~\citep{wang2024sibyl} on the GAIA benchmark. GAIA encompasses tasks from various domains such as web browsing, file reading, and multimodal understanding, making it challenging to pursue a single optimal multi-agent system for all tasks. Thus, the modest improvements of AFlow and ADAS over vanilla LLMs (only $3.35\%\uparrow$ and $2.04\%\uparrow$ on average) are understandable. In contrast, \ourmethod can adaptively sample customized agentic systems for different domains, achieving $18.38\%$ and $17.61\%$ improvements on Level 1 and 2 tasks, respectively.

% 可以观察到， ADAS and AFlow的性能提升相较于vanilla LLM 极为有限，我们归结于其力图优化单个最优workflow的范式与GAIA benchmark本身起了冲突：GAIA包括web

% It is worth noting that AFlow's performance here is slightly lower than that reported in its original paper~\citep{zhang2024aflow}, as the original experiments utilized the more advanced \llmname{claude-3.5-sonnet}, whereas for fair comparison, our experiments consistently used \llmname{gpt-4o-mini}.

\begin{figure}[!t]
\centering
\includegraphics[width=1.0\linewidth]{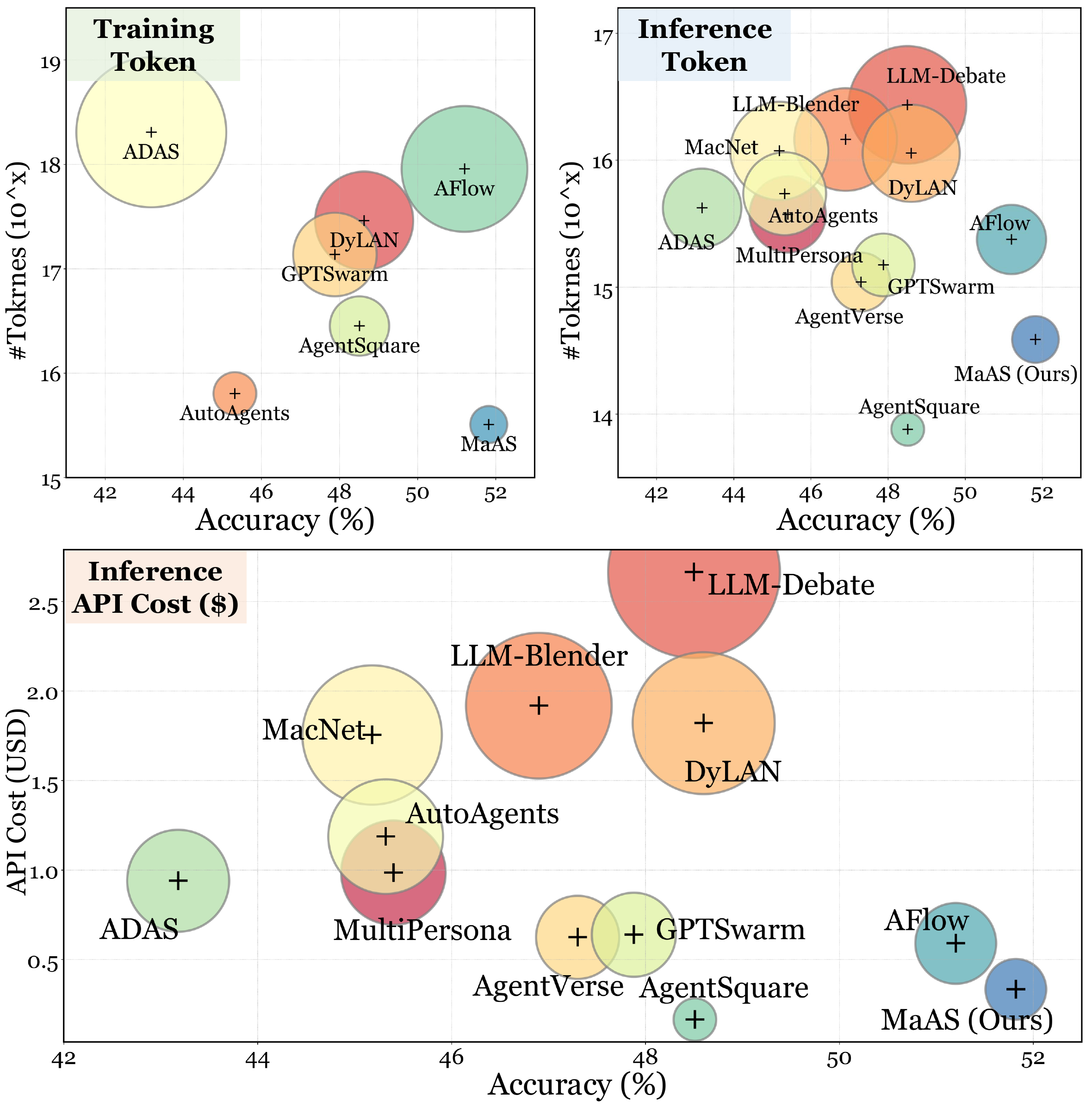}
\vspace{-2em}
\caption{The cost analysis of \ourmethod on MATH benchmark.
}
\vspace{-1.4em}
\label{fig:rq2-cost}
\end{figure}

\begin{figure}[!t]
\centering
\includegraphics[width=1.0\linewidth]{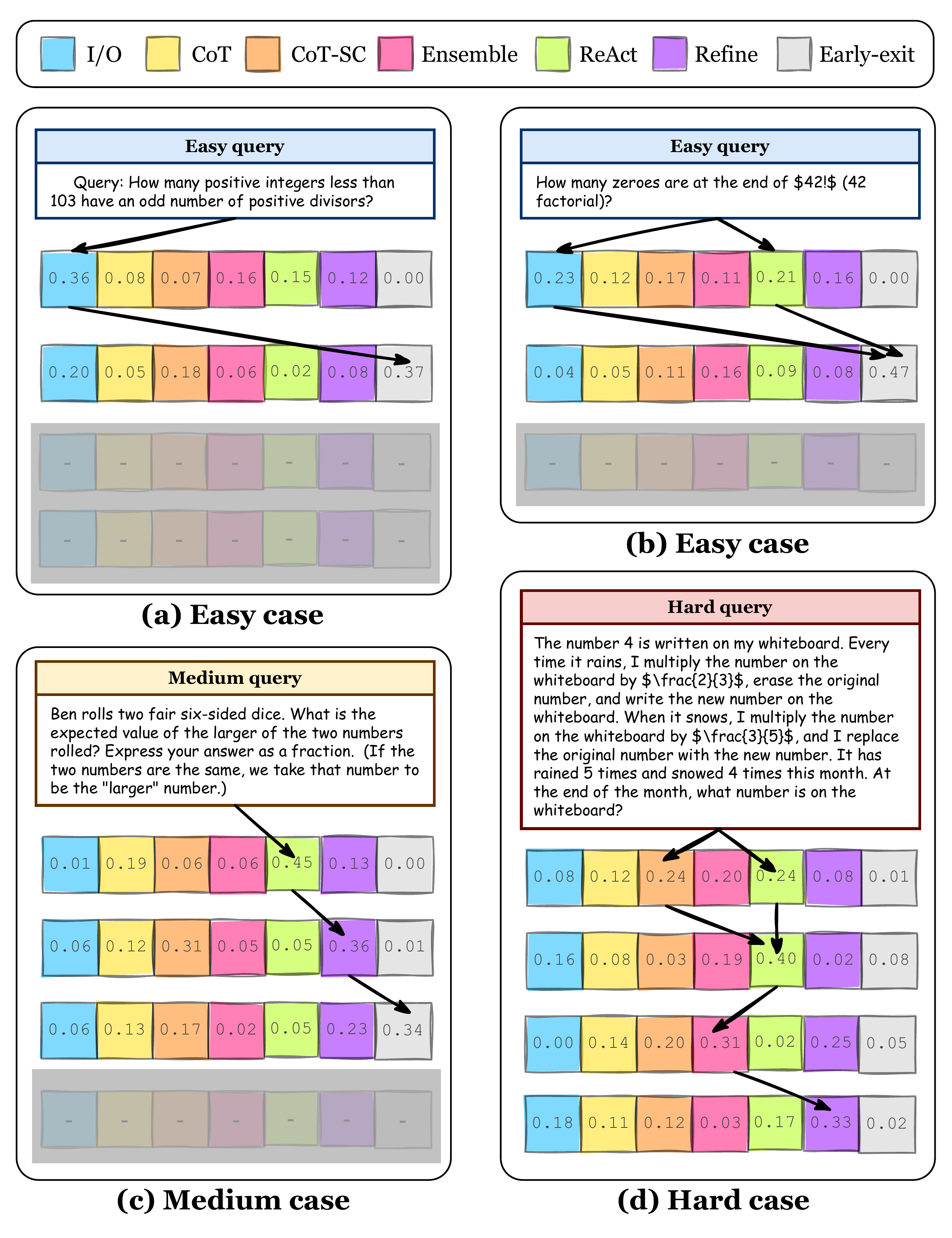}
\vspace{-2.5em}
\caption{The visualization of \ourmethod's operator sampling process.
}
\vspace{-1.5em}
\label{fig:probability}
\end{figure}

\begin{figure*}[t]
\centering
\includegraphics[width=1.0\linewidth]{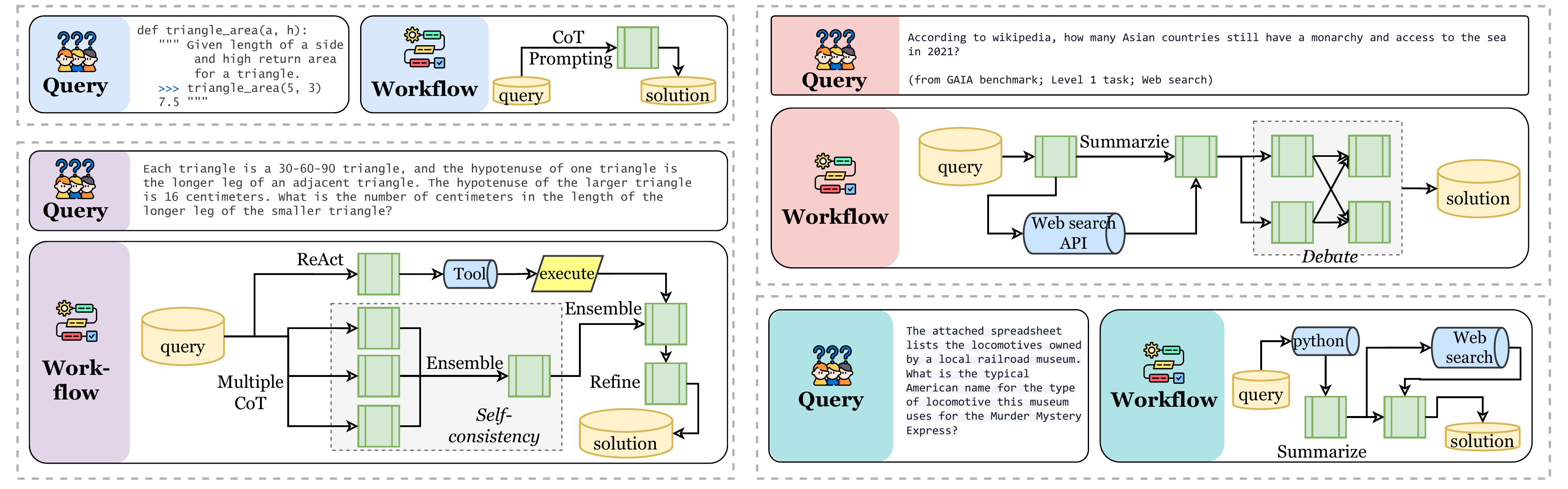}
\vspace{-2em}
\caption{Case study and visualization for \ourmethod. Queries are from HumanEval, MATH and GAIA benchmarks.
}
\vspace{-0.5em}
\label{fig:case}
\end{figure*}

\vspace{-0.3em}
\subsection{Cost Analysis}
\vspace{-0.4em}
To answer RQ2, we demonstrate that \textbf{\ourmethod is both training/inference cost-efficient} from the following three dimensions: (1) token cost, (2) API cost, and (3) wall-clock time, as shown in \Cref{tab:efficiency} and \Cref{fig:rq2-cost}. We observe:

\textbf{Obs.\ding{183} \ourmethod's optimization is resource-friendly.} As shown in \Cref{fig:rq2-cost} \textit{(Training Tokens)}, among the various optimization-oriented agentic workflows, \ourmethod achieves the highest accuracy with the least training token consumption. While AFlow's accuracy is comparable to that of \ourmethod, its training cost reaches $22.50\$$, which is $6.8\times$ that of \ourmethod (merely $3.38\$$). Additionally, existing agentic automation pipelines are relatively time-consuming, with DyLAN taking $508$ minutes and GPTSwarm taking $129$ minutes. In contrast, the optimization wall-clock time of \ourmethod requires only $53$ minutes.

\textbf{Obs.\ding{184} Agentic supernet enjoys superior token economy during inference.} As shown in \Cref{fig:rq2-cost} \textit{(Inference API Cost)}, \ourmethod achieves the highest accuracy with an API cost of $0.42\$$, demonstrating its high performance and token economy. Although AgentSquare's API cost is slightly lower than \ourmethod's, this is due to its limitation to a single-agent search, which severely restricts its performance (resulting in a $4\%$ drop compared to \ourmethod). \Cref{tab:efficiency} further highlights that \ourmethod has the lowest prompt/completion token consumption, the lowest API cost, and the shortest wall-clock time during inference. These advantages can be attributed to the agentic supernet's ability to dynamically allocate resources based on the difficulty of the query.

\vspace{-0.4em}
\subsection{Case Study}
\vspace{-0.4em}
In this section, we explore and visualize the intrinsic mechanisms of the agentic supernet. \Cref{fig:probability} showcases the probability distributions of the agentic supernet when faced with different queries, and \Cref{fig:case} presents the multi-agent systems designed by \ourmethod for queries from the MATH, GAIA, and HumanEval benchmarks. We have:

\textbf{Obs.\ding{185} \ourmethod learns to query-aware early exit from the reasoning process.} As shown in \Cref{fig:probability}, when faced with the easy queries (a) and (b), \ourmethod exits multi-agent architecture sampling at the second layer with probabilities of $0.37$ and $0.47$, respectively, selecting the early-exit operator. Notably, query (b) chose two agentic operators at the first layer: direct I/O and ReAct, demonstrating \ourmethod's ability to dynamically allocate different operators at each layer (corresponding to \Cref{eq:layer-l}). For the more challenging queries (c) and (d), \ourmethod sampled additional layers, further proving its ability to customize the multi-agent system based on query awareness. This is also visualized by \Cref{fig:dynamic}, in which the probability of $\mathcal{O}_\text{exit}$ becomes increasingly high with the supernet depth increases.

\begin{figure}[!t]
\centering
\includegraphics[width=1.0\linewidth]{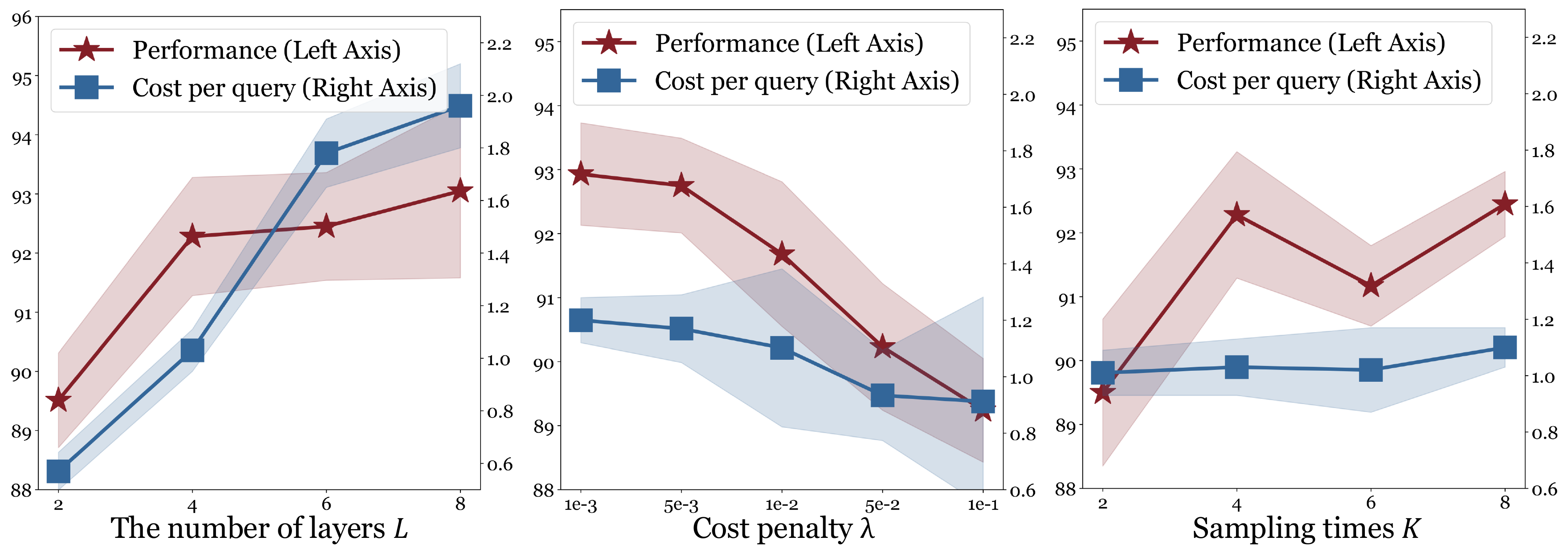}
\vspace{-2em}
\caption{Parameter sensitivity analysis of \ourmethod. The unit of cost per query (right) and performance (left) is $10^{-3}\cdot\$$ and $pass@1~(\%)$, respectively.
}
\vspace{-0.9em}
\label{fig:sensi}
\end{figure}

\vspace{-0.6em}
\subsection{Framework Analysis}
\vspace{-0.4em}
\paragraph{Sensitivity Analysis} We analyze the sensitivity of \ourmethod to three core parameters: the number of layers in the agentic supernet $L$, the cost penalty coefficient $\lambda$ in \Cref{eq:objective}, and the sampling count $K$ in \Cref{eq:loss-pi}. The results are presented in \Cref{fig:sensi}. 
\textbf{For the parameter $L$}, we observe a significant performance improvement as $L$ increases from 2 to 4 ($89.5\% \rightarrow 92.8\%$). However, further increases yield only marginal performance gains while incurring higher per-query inference costs. Considering both performance and cost, we select $L = 4$. 
\textbf{For the parameter $\lambda$}, we find that larger values lead \ourmethod to favor more cost-efficient solutions, albeit with some performance degradation. 
\textbf{For the parameter $K$}, we note that performance is suboptimal with highest variance when $K = 2$. Increasing $K$ to 4 effectively achieves a satisfactory low-variance estimation.

\vspace{-0.9em}
\paragraph{Ablation Study} We perform an ablation study on three key components of \ourmethod: \textbf{(1) \textit{w/o} $\nabla_\mathbb{O}\mathcal{L}$}, removing the textual gradient in \Cref{eq:loss-operator}; \textbf{(2) \textit{w/o} $\mathbb{O}_\text{exit}$}, removing the early-exit operator in \Cref{eq:early-exit}; and \textbf{(3) \textit{w/o} $C(\cdot)$}, eliminating the cost constraint in \Cref{eq:objective}. 
We observe from \Cref{tab:ablation} that removing the textual gradient causes the largest performance drop, as it disables \ourmethod's self-evolving capability. Removing $\mathbb{O}_\text{exit}$ and $C(\cdot)$ results in little impact on performance, but it weakens \ourmethod's query-dependent nature and unnecessarily increases the inference cost.

\begin{table}[!htpb]
\vspace{-1em}
\caption{Ablation study of \ourmethod. }\label{tab:ablation}
% \begin{center}
\vspace{0.1em}
\centering
\resizebox{\columnwidth}{!}{
\begin{tabular}{c|cc|cc}
\toprule
\makecell{Dataset} &\multicolumn{2}{c|}{HumanEval} & \multicolumn{2}{c}{MATH}\\
\midrule
\makecell{Metric}  &  \makecell{Pass@1 \\(\%)} & \makecell{Cost\\ ($10^{-3}\;\$$)}   &  \makecell{Accuracy \\(\%)} & \makecell{Cost\\ ($10^{-3}\;\$$)}  \\
\midrule
\makecell{Vanilla \ourmethod}    & $92.85$ & $1.01$ & $51.82$ & $0.86$ \\
\midrule
\ourmethod \textit{w/o} $\nabla_\mathbb{O}\mathcal{L}$  & $90.17$ & $0.90$ & $48.23$ & $0.84$ \\
\ourmethod \textit{w/o} $\mathbb{O}_\text{exit}$ & $91.44$ & $1.67$ &  $51.53$ & $1.04$   \\
\ourmethod \textit{w/o} $C(\cdot)$ & $92.94$ & $1.38$ & $51.19$ & $1.28$\\
\bottomrule
\end{tabular}}
% \end{center}
\vspace{-0.2em}
\end{table}
\vspace{-0.4em}
% \paragraph{Inductive Ability}

\vspace{-0.4em}
\paragraph{Transferability Analysis.} We evaluate whether the agentic supernet of \ourmethod is \textbf{(1) model-agnostic} and \textbf{(2) generalizable across datasets}, with results presented in \Cref{tab:cross-llm,tab:cross-dataset}. As shown, the agentic supernet optimized by \ourmethod transfers well to models such as \llmname{Qwen-2.5-70b}, with $4.98\%\sim5.50\%\uparrow$ in performance, while also demonstrating strong cross-dataset generalization.
\vspace{-0.4em}
\paragraph{Inductive Analysis.} To evaluate whether \ourmethod possesses inductive capabilities, \textit{i.e.}, the ability to generalize to unseen agentic operators, we select the Debate~\cite{arXiv2023_MultiAgent-Debate} operator as a holdout. We then compare the operator distribution of \ourmethod during inference with and without Debate in \Cref{fig:dynamic,fig:dynamic2}. The results demonstrate that \ourmethod can still reasonably activate and utilize the unseen operator at an appropriate proportion.

\vspace{-0.4em}
\section{Conclusion}
\vspace{-0.4em}
In this paper, we \textit{for the first time} shift the paradigm of automated multi-agent system design from seeking a (possibly non-existent) single optimal system to optimizing a probabilistic, continuous distribution of agentic architectures, termed the \textbf{agentic supernet}. Building on this concept, we propose \ourmethod, which dynamically samples multi-agent systems that deliver satisfactory performance and token efficiency for user queries across different domains and varying levels of difficulty. %Extensive evaluations and visualizations demonstrate that \ourmethod autonomously searches for the most performant multi-agent systems across six benchmarks, consuming minimal training and inference resources. 
We believe that \ourmethod paves the way toward fully automated, self-organizing, and self-evolving collective intelligence.

% \subsection{Theorems and such}
% The preferred way is to number definitions, propositions, lemmas, etc. consecutively, within sections, as shown below.
% \begin{definition}
% \label{def:inj}
% A function $f:X \to Y$ is injective if for any $x,y\in X$ different, $f(x)\ne f(y)$.
% \end{definition}
% Using \cref{def:inj} we immediate get the following result:
% \begin{proposition}
% If $f$ is injective mapping a set $X$ to another set $Y$, 
% the cardinality of $Y$ is at least as large as that of $X$
% \end{proposition}
% \begin{proof} 
% Left as an exercise to the reader. 
% \end{proof}
% \cref{lem:usefullemma} stated next will prove to be useful.
% \begin{lemma}
% \label{lem:usefullemma}
% For any $f:X \to Y$ and $g:Y\to Z$ injective functions, $f \circ g$ is injective.
% \end{lemma}
% \begin{theorem}
% \label{thm:bigtheorem}
% If $f:X\to Y$ is bijective, the cardinality of $X$ and $Y$ are the same.
% \end{theorem}
% An easy corollary of \cref{thm:bigtheorem} is the following:
% \begin{corollary}
% If $f:X\to Y$ is bijective, 
% the cardinality of $X$ is at least as large as that of $Y$.
% \end{corollary}
% \begin{assumption}
% The set $X$ is finite.
% \label{ass:xfinite}
% \end{assumption}
% \begin{remark}
% According to some, it is only the finite case (cf. \cref{ass:xfinite}) that is interesting.
% \end{remark}
%restatable

\section*{Acknowledgements}

This research is supported by the National Natural Science Foundation of China (No. 92270114), and the Shanghai Municipal Science and Technology Major Project.

\section*{Impact Statement}  

\paragraph{Ethical Considerations.}  
We believe that our proposed \ourmethod framework raises no ethical concerns regarding its motivation, design, implementation, or data usage. The method is designed to advance the automation of multi-agent systems in a principled and resource-efficient manner, ensuring responsible development while adhering to ethical guidelines in AI research.
\vspace{-0.5em}  
\paragraph{Societal Implications.}  
\ourmethod introduces a new paradigm in multi-agent system design by replacing static, one-size-fits-all architectures with a dynamic and adaptive agentic supernet. This approach enables fine-grained resource allocation tailored to query difficulty and domain, significantly improving efficiency while maintaining high-quality outputs. By reducing inference costs and enhancing the flexibility of multi-agent workflows, \ourmethod has the potential to democratize access to intelligent automation across diverse applications, including education, research, and industry. 


\bibliography{example_paper}
\bibliographystyle{icml2025}

%%%%%%%%%%%%%%%%%%%%%%%%%%%%%%%%%%%%%%%%%%%%%%%%%%%%%%%%%%%%%%%%%%%%%%%%%%%%%%%
%%%%%%%%%%%%%%%%%%%%%%%%%%%%%%%%%%%%%%%%%%%%%%%%%%%%%%%%%%%%%%%%%%%%%%%%%%%%%%%
% APPENDIX
%%%%%%%%%%%%%%%%%%%%%%%%%%%%%%%%%%%%%%%%%%%%%%%%%%%%%%%%%%%%%%%%%%%%%%%%%%%%%%%
%%%%%%%%%%%%%%%%%%%%%%%%%%%%%%%%%%%%%%%%%%%%%%%%%%%%%%%%%%%%%%%%%%%%%%%%%%%%%%%
\newpage
\appendix
\onecolumn

\section{Notations}

\begin{table}[!h]
\centering
\caption{Notations and Definitions}
\begin{tabular}{lp{12cm}}
\Xhline{1.2pt}
\textbf{Notation} & \textbf{Definition} \\ 
\Xhline{1.pt}
$\mathcal{O} = \{\{\mathcal{M}_i\}_{i=1}^m, \mathcal{P}, \{\mathcal{T}_i\}_{i=1}^n\}$ & An agentic operator comprising a set of LLM instances, a textual prompt, and a set of temperature settings. \\ 
% \midrule
$\mathcal{M}$ & An individual LLM instance. \\ 
% \midrule
$\mathbb{M}$ & The set of all feasible LLMs. \\ 
% \midrule
$\mathcal{P}$ & A textual prompt used as input to the LLM. \\ 
% \midrule
$\mathbb{P}$ & The feasible space of prompts. \\ 
% \midrule
$\mathcal{T}$ & The temperature setting of the LLM. \\ 
% \midrule
$\mathbb{O}$ & The set of all feasible agentic operators. \\ 
% \midrule
$\mathcal{G} = \{\mathcal{V},\mathcal{E}\}$ & A multi-agent system represented as a graph with vertices $\mathcal{V}$ and edges $\mathcal{E}$. \\ 
% % \midrule
$\mathcal{A} = \{\boldsymbol{\pi}, \mathbb{O}\} = \{\pi_{\ell}(\mathcal{O})\}_{\mathcal{O} \in \mathbb{O}}\}_{\ell=1}^L$ & An $L$-layer probabilistic agentic supernet, consisting of a distribution $\boldsymbol{\pi}$ and a set of feasible operators $\mathbb{O}$. \\ 
% \midrule
$\boldsymbol{\pi}$ & The distribution associated with the agentic supernet. \\ 
% \midrule
$U(\mathcal{G}; q,a)$ & The utility evaluator of $\mathcal{G}$ with respect to query $q$ and answer $a$. \\ 
% \midrule
$C(\mathcal{G}; q,a)$ & The cost evaluator of $\mathcal{G}$ with respect to query $q$ and answer $a$. \\ 
% \midrule
$\mathbb{Q}_\phi$ & The controller network parameterized by $\phi$. \\ 
% \midrule
$e(a\|\mathcal{G})$ & Execution of $\mathcal{G}$ to produce the answer $a$. \\ 
% \midrule
$\mathcal{V}_\ell$ & The selected operators at layer $\ell$ of the agentic supernet $\mathcal{A}$. \\ 
% \midrule
$\mathcal{O}_\text{exit}$ & The early-exit operator. \\ 
% \midrule
$\mathbf{v}(\cdot)$ & The text embedding function. \\ 
% \midrule
$\nabla_\pi\mathcal{L}$ & The gradient of the loss $\mathcal{L}$ with respect to the distribution $\boldsymbol{\pi}$. \\ 
% \midrule
$\nabla_\mathbb{O}\mathcal{L}$ & The textual gradient of the loss $\mathcal{L}$ with respect to the operators $\mathbb{O}$. \\ 
\Xhline{1.2pt}
\end{tabular}
\label{tab:notations}
\end{table}

\section{Technical Details}
\subsection{Operator Space}\label{app:operator}

In this section, we detail the initialization of operator nodes as follows:

\begin{enumerate}
    \item \textbf{Chain-of-Thought (CoT).}  
    CoT~\citep{cot} reasoning encourages the LLM to think step by step rather than directly outputting an answer. This approach enhances its capability to solve complex problems through intermediate reasoning steps, improving task handling and providing greater transparency in the decision-making process.

    \item \textbf{LLM-Debate.}  
    LLM-Debate~\citep{arXiv2023_MultiAgent-Debate} allows multiple LLMs to debate, leveraging diverse perspectives to identify better solutions. In practice, we initialize three debaters and permit up to two debate rounds.

    % \item \textbf{Take a Step Back.}  
    % As proposed by \citet{zheng2023take-a-step-back}, this operator prompts the LLM to first consider the principles underlying the task. By focusing on foundational principles, the model enhances its reasoning and delivers more accurate solutions.

    \item \textbf{Self-Consistency.}  
    Adopting the methodology from \citet{wang2023selfconsistency}, this operator aggregates five CoT reasoning paths and determines the final answer through majority voting.

    \item \textbf{Self-Refine.}  
    Following \citet{NeurIPS2023_Self-Refine}, this operator initially generates an answer using CoT reasoning, then prompts the agent to self-reflect iteratively. We set a maximum of five refinement iterations.

    \item \textbf{Ensemble.}  
    Inspired by LLM-Blender~\citep{blender}, this operator involves three LLM-powered agents from different sources outputting answers to the same query. The pairwise ranking is used to evaluate and aggregate their responses into a final solution.

    \item \textbf{Testing.} Following the test designer in AgentCoder~\cite{huang2023agentcoder}, this operator is used for generating test cases for the generated code.

    \item \textbf{ReAct.} Following~\citep{yao2023react}, this operator enables the agent to leverage versatile tools, including code interpreter, web searching, external knowledge database, \textit{etc.}, to handle diverse user demands. 

    \item \textbf{Early exit.} We introduce the early exit operator, which interrupts the multi-agent architecture sampling process and enables the depth of the agentic supernet to be query-dependent.

\end{enumerate}

We respectfully note that the selection of these operators is highly customizable, allowing users the flexibility to incorporate their desired operators into the operator repository of \ourmethod.

\subsection{Embedding Function}\label{app:embedding}

Following established practices~\cite{feng2024graphrouter}, we first employ an LLM to generate a comprehensive profile description for each operator. Subsequently, a lightweight text embedding model (in our case, MiniLM~\citep{wang2020minilm}) is used to encode the profile into a fixed-dimensional embedding. The prompt for generating the operator profile is as follows:

\begin{tcolorbox}[notitle, sharp corners, breakable, colframe=Periwinkle, colback=white, 
       boxrule=3pt, boxsep=0.5pt, enhanced, 
       shadow={3pt}{-3pt}{0pt}{opacity=1,mygrey},
       title={Embedding Prompt},]\label{box:operator-profile}
       \footnotesize
       {\fontfamily{pcr}\selectfont
\begin{lstlisting}
prompt = """You are a highly proficient expert in designing and defining operators for large language models (LLMs). Your primary objective is to meticulously generate the `description` and `interface` fields for a specified operator based on its provided Python implementation. The generated content must be accurate, efficient, and precisely reflect the functionality of the operator's code.

To ensure consistency, quality, and adherence to best practices, refer to the following examples of previously defined operators:
{
    "Generate": {
        "description": "Generates anything based on customized input and instruction.",
        "interface": "generate(input: str, instruction: str) -> dict with key 'response' of type str" 
    },
    "ScEnsemble": {
        "description": "Uses self-consistency to select the solution that appears most frequently in the solution list, improving the selection to enhance the choice of the best solution.",
        "interface": "sc_ensemble(solutions: List[str], problem: str) -> dict with key 'response' of type str"
    }
}

Now, given the following operator code. This code encompasses the function signature, parameters with type annotations, internal logic, and return statements essential for comprehensively understanding the operator's purpose and behavior.Please provide its `description` and `interface` fields in the same format.
[operator code]

 """
\end{lstlisting}
}
\end{tcolorbox}

\subsection{Textaul Gradient}\label{app:text-gradient}

The implementation of the textual gradient component is partially adapted from the repositories \url{https://github.com/ShengranHu/ADAS/} and \url{https://github.com/tsinghua-fib-lab/agentsquare}. We would like to explicitly acknowledge this contribution and express our sincere gratitude to the authors for their open-source efforts.

\begin{tcolorbox}[notitle, sharp corners, breakable, colframe=Periwinkle, colback=white, 
       boxrule=3pt, boxsep=0.5pt, enhanced, 
       shadow={3pt}{-3pt}{0pt}{opacity=1,mygrey},
       title={Textual Gradient},]\label{box:gradient}
       \footnotesize
       {\fontfamily{pcr}\selectfont
\begin{lstlisting}
base = """
# Overview
You are an expert machine learning researcher specializing in designing agentic systems. Your objective is to create building blocks such as prompts and control flows within these systems to solve complex tasks. Specifically, you aim to design an optimal agent that performs exceptionally on the HumanEval benchmark. The HumanEval dataset evaluates code generation capabilities in AI systems, consisting of 164 hand-crafted Python programming problems. Each problem includes: - A function signature with a docstring describing the task - Test cases to verify functional correctness

# Example Question from HumanEval
[An example question from HumanEval dataset here]

# Operator code template:
class Operator:
    def __init__(self, llm: LLM, name: str):
        self.name = name
        self.llm = llm

    def __call__(self, *args, **kwargs):
        raise NotImplementedError

    async def _fill_node(self, op_class, prompt, mode=None, **extra_kwargs):
        fill_kwargs = {"context": prompt, "llm": self.llm}
        if mode:
            fill_kwargs["mode"] = mode
        fill_kwargs.update(extra_kwargs)
        node = await ActionNode.from_pydantic(op_class).fill(**fill_kwargs)
        return node.instruct_content.model_dump()

class GenerateOp(BaseModel):
    response: str = Field(default="", description="Your solution for this problem")

class Generate(Operator):
    GENERATE_PROMPT = '''
You are tasked with solving the following Python programming problem. Generate a complete, syntactically correct Python function that strictly adheres to the given requirements.

Problem:
{input}

Follow these steps:
1. Analyze the problem requirements and identify edge cases
2. Design a solution that passes all implied test cases
3. Implement the function with clear variable names and comments

Ensure:
- The code directly implements the requested functionality
- All parameters and return types match the problem specification
- Exception handling for edge cases is included when necessary '''

    def __init__(self, llm: LLM, name: str = "Generate"):
        super().__init__(llm, name)

    async def __call__(self, input: str, mode: str = None):
        prompt = self.GENERATE_PROMPT.format(input=input)
        response = await self._fill_node(GenerateOp, prompt, mode="xml_fill")
        return response

# Discovered architecture archive
Here is the archive of the discovered operator architectures:
[ARCHIVE]

# Output Instruction and Example:
The output should be a JSON object with the following structure.The first key should be ("thought"), and it should capture your thought process for designing the next operator. The second key ("description") corresponds to the brief description of your next operator. Finally, the last key ("code") corresponds to the exact operator and its prompt in Python code that you would like to try. You must write COMPLETE CODE in "code": Your code will be part of the entire project, so please implement complete, reliable, reusable code snippets.

- thought: Captures your thought process for designing the next operator.
  - Reason about what the next interesting operator should be.
  - Describe your reasoning and the overall concept behind the operator design.
  - Detail the implementation steps.
- description: A brief description of your next operator.
- code: The exact operator and its prompt in Python code. Ensure the code is complete, reliable, and reusable.


Here is an example of the output format for the next operator: 
[operator_example]

You must strictly follow the exact input/output interface used above. Also, it could be helpful to set the LLM's role and temperature to further control the LLM's response. DON'T try to use some function that doesn't exist. In __call__(), you need to specify the instruction, input information, the prompt and the required output fields class for operators to do their specific part of the architecture. 

# Your task 
You are highly proficient in prompting techniques and well-versed with agentic systems from academic literature. Your goal is to maximize performance metrics by proposing innovative and effective new operators.
Instructions:
1. Analyze the Discovered Operators: Carefully review the operators in the archive to identify strengths, weaknesses, and areas for improvement.
2. Draw Insights: Extract lessons and insights from existing operators to inform the design of the next operator.
3. Innovate: Think creatively to design an operator that addresses current limitations or explores new functionalities, drawing inspiration from related agent papers or other research areas.
4. Design the Operator: Propose the next operator's `thought`, `description`, and `code` following the specified format.
5. Ensure Completeness: The generated code must be complete, reliable, and reusable, fitting seamlessly into the existing architecture.

Execution Steps:
1. Insert Operator Code: Replace the `[ARCHIVE]` and `[operator_example]` placeholders with actual content as needed.
2. Generate Output: Produce the `thought`, `description`, and `code` fields for the new operator, ensuring adherence to the guidelines.
3. Validate Output: Ensure the generated JSON is correctly formatted and the code is syntactically and functionally correct.

THINK OUTSIDE THE BOX and leverage interdisciplinary insights to enhance the agentic system's capabilities.
"""
\end{lstlisting}
}
\end{tcolorbox}

%%%%%%%%%%%%%%%%%%%%%%%%%%%%%%%%%%%%%%%%%%%%%%%%%%%%%%%%%%%%%%%%%%%%%%%%%%%%%%%

\section{Experimental Details}
\subsection{Dataset Statistics}\label{app:dataset}

Building upon established methodologies in workflow automation~\cite{saad2024archon,hu2024adas,zhang2024aflow}, we divide each dataset into training and test sets using a \textsc{train:test} ratio of 1:4. For the MATH benchmark, we adhere to \cite{hong2024datainterpreter}, selecting a subset of 617 harder problems spanning four representative categories, Combinatorics \& Probability, Number Theory, Pre-algebra, and Pre-calculus, all at difficulty level 5. The dataset statistics are included in \Cref{tab:dataset}.

\begin{table}[!h]
\vspace{-1em}
\caption{Dataset Statistics.}\label{tab:dataset}
% \begin{center}
\vspace{0.1em}
\centering
\begin{tabular}{l|cccc}
\toprule
Domain & Dataset & \#Train & \#Test & Metric  \\ 
\midrule
\multirow{2}{*}{Code Generation} & HumanEval & 33 & 131 & pass@1\\ 
& MBPP & 86 & 341 & pass@1\\ 
\midrule
\multirow{3}{*}{Math Reasoning}& GSM8K & 264 & 1055 & Accuracy\\ 
& MATH & 119 & 486 & Accuracy\\ 
& MultiArith &150 & 600 & Accuracy\\
\midrule
Tool use & GAIA & 94 & 372 & Accuracy\\
\bottomrule
\end{tabular}
\end{table}

\subsection{Baseline Setups}\label{app:baselines}
%%%%%%%%%%%%%%%%%%%%%%%%%%%%%%%%%%%%%%%%%%%%%%%%%%%%%%%%%%%%%%%%%%%%%%%%%%%%%%%
%%%%%%%%%%%%%%%%%%%%%%%%%%%%%%%%%%%%%%%%%%%%%%%%%%%%%%%%%%%%%%%%%%%%%%%%%%%%%%%

In this section, we provide a detailed description of the configurations for baseline methods:  

\begin{enumerate}  
    \item \textbf{CoT.} Chain-of-Thought (CoT) prompting guides LLM agents to break down reasoning into sequential steps rather than generating direct answers. We employ the implementation from \cite{zhang2022automaticcot}.  

    \item \textbf{ComplexCoT.} We follow the official implementation available at \url{https://github.com/FranxYao/Complexity-Based-Prompting/tree/main}.  

    \item \textbf{Self-consistency.} To enhance robustness, we aggregate five CoT-generated solutions.  

    \item \textbf{LLM-Debate.} We instantiate five LLM-agents, each assigned a distinct role, which participate in up to two rounds of debate, after which the final decision is determined via majority voting. The implementation is based on \url{https://github.com/ucl-dark/llm_debate}.  

    \item \textbf{LLM-Blender.} We choose two \llmname{gpt-4o-mini}, one \llmname{Qwen-2.5-72b}, and one \llmname{llama-3.1-70b} to empower LLM-Blender~\cite{blender}.  

    \item \textbf{DyLAN.} We directly utilize the implementation from \cite{arXiv2023_Dynamic-LLM-Agent}.  

    \item \textbf{AgentVerse.} The experimental setup follows the original implementation from \cite{chen2023agentverse}.  

    \item \textbf{MacNet.} For MacNet~\citep{qian2024scaling}, we adopt the ``MacNet-MESH'' variant, which corresponds to a fully connected network topology.  

    \item \textbf{GPTSwarm.} The method is implemented in accordance with the original settings described in \cite{zhuge2024gptswarm}.  

    \item \textbf{AutoAgents.} We adhere to the official configuration specified in \cite{chen2023autoagents}.  

    \item \textbf{ADAS.} The implementation details are directly inherited from \cite{hu2024adas}.  

    \item \textbf{AgentSquare.} We utilize the modular search framework introduced in \cite{shang2024agentsquare}. The base LLM remains fixed at \llmname{gpt-4o-mini}, with early stopping set to a patience of 5 iterations.  

    \item \textbf{AFlow.} In \cite{zhang2024aflow}, AFlow operates with both \llmname{gpt-4o-mini} and \llmname{claude-3.5-sonnet}. To maintain fairness under homogeneous conditions, we restrict AFlow to \llmname{gpt-4o-mini} and set \textsc{max\_iteration}=20.  
\end{enumerate}

\section{Supplementary Results}
\label{app:exp-result}

\begin{figure*}[t]
\centering
\includegraphics[width=\linewidth]{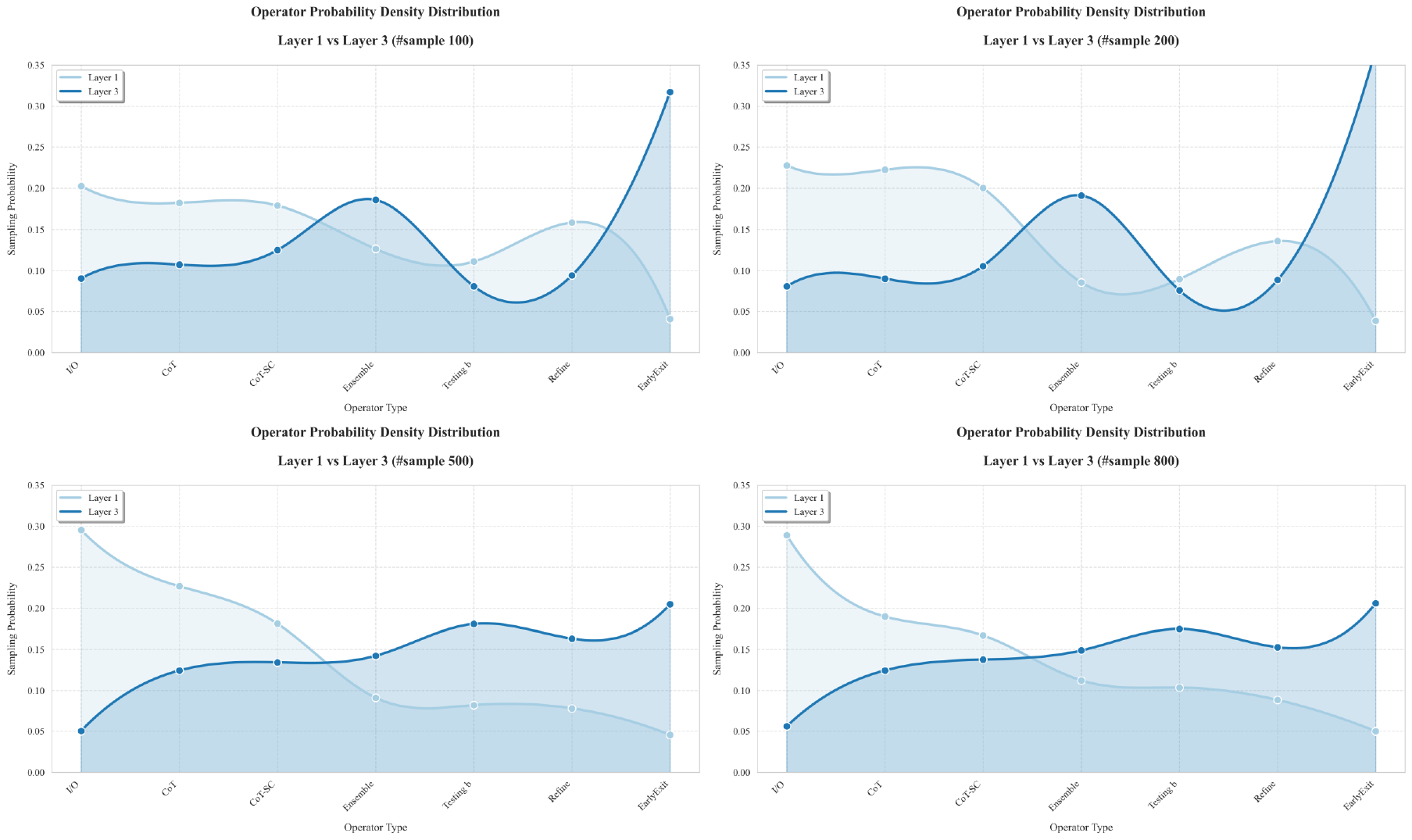}
\vspace{-1em}
\caption{The evolution of operator sampling trends as the sampling count increases.
}
\vspace{-0.5em}
\label{fig:distribution}
\end{figure*}

We have visualized the evolution of operator sampling trends as the sampling count increases, in \Cref{fig:distribution}. \ourmethod learns to avoid overly confident early stopping and instead prioritizes testing and self-refinement in deeper layers.

\begin{table}[!htpb]
\vspace{-1em}
\caption{Cross-model transferability of \ourmethod. We optimize the agentic supernet with \llmname{gpt-4o-mini}, and report the performances before and after equipping the LLM backbones with the optimized agentic supernet.}\label{tab:cross-llm}
% \begin{center}
\vspace{0.1em}
\centering
% \resizebox{\columnwidth}{!}{
\begin{tabular}{c|ccc}
\toprule
\makecell{Dataset} &\multicolumn{3}{c}{HumanEval}\\
\midrule
\makecell{LLM Backbone}  &  \llmname{gpt-4o-mini} & \llmname{Qwen-2.5-72b} & \llmname{llama-3.1-70b}   \\
\midrule
vanilla    & $87.08$ & $85.60$ & $80.06$ \\
+\ourmethod    & $92.85$ & $90.14$ & $85.26$ \\

\midrule

\makecell{Dataset} &\multicolumn{3}{c}{MATH}\\
\midrule
\makecell{LLM Backbone}  &  \llmname{gpt-4o-mini} & \llmname{Qwen-2.5-70b}  & \llmname{llama-3.1-70b}  \\
\midrule
vanilla    & $46.29$ & $63.80$ & $31.93$ \\
+\ourmethod    & $51.82$ & $69.35$ & $42.97$ \\% \midrule
\bottomrule
\end{tabular}
%}
% \end{center}
\vspace{-0.2em}
\end{table}
% \vspace{-0.4em}

\begin{table}[!htpb]
\vspace{-1em}
\caption{Cross-dataset transferability of \ourmethod. ``MATH$\rightarrow$GSM8K'' denotes optimizing the agentic supernet on MATH and evaluating it on GSM8K, with similar notation applied to other cases.}\label{tab:cross-dataset}
% \begin{center}
\vspace{0.1em}
\centering
% \resizebox{\columnwidth}{!}{
\begin{tabular}{c|ccc}
\toprule
%\makecell{Dataset} &\multicolumn{3}{c}{HumanEval}\\
% \midrule
\makecell{Transfer}  & MATH$\rightarrow$GSM8K  & GSM8K$\rightarrow$MATH &  HumanEval$\rightarrow$ MATH \\
\midrule
GPTSwarm    & $89.96$ & $45.18$ & $47.92$ \\
AFlow    & $91.95$ & $49.39$ & $47.15$ \\
\ourmethod & $92.80$ & $51.02$ & $50.27$\\

\bottomrule
\end{tabular}
%}
% \end{center}
\vspace{-0.2em}
\end{table}
% \vspace{-0.4em}

\begin{figure*}[t]
\centering
\includegraphics[width=0.85\linewidth]{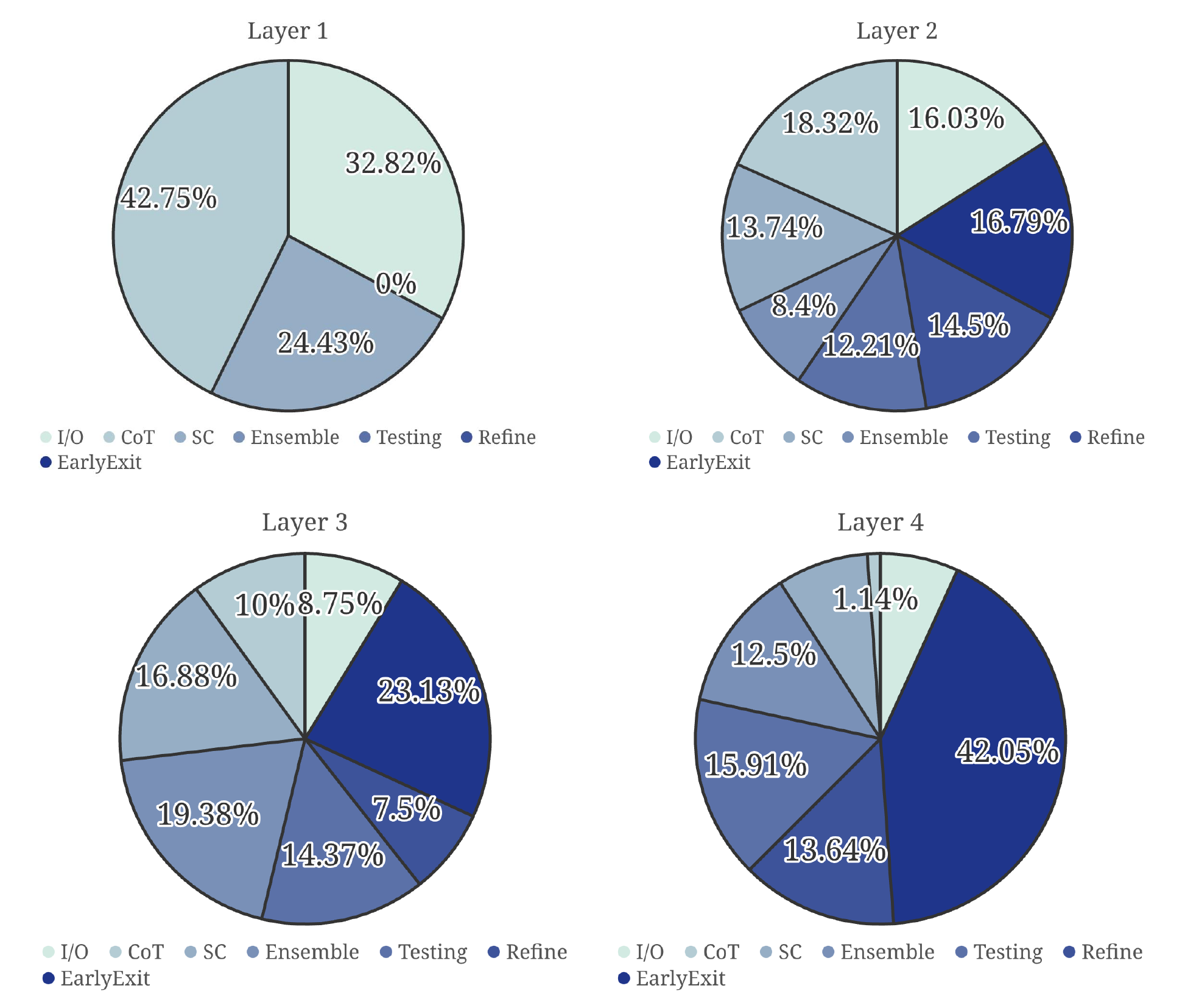}
\vspace{-1em}
\caption{The layer-wise distribution of \ourmethod on HumanEval benchmark without Debate operator.
}
\vspace{-0.5em}
\label{fig:dynamic}
\end{figure*}

\begin{figure*}[t]
\centering
\includegraphics[width=0.85\linewidth]{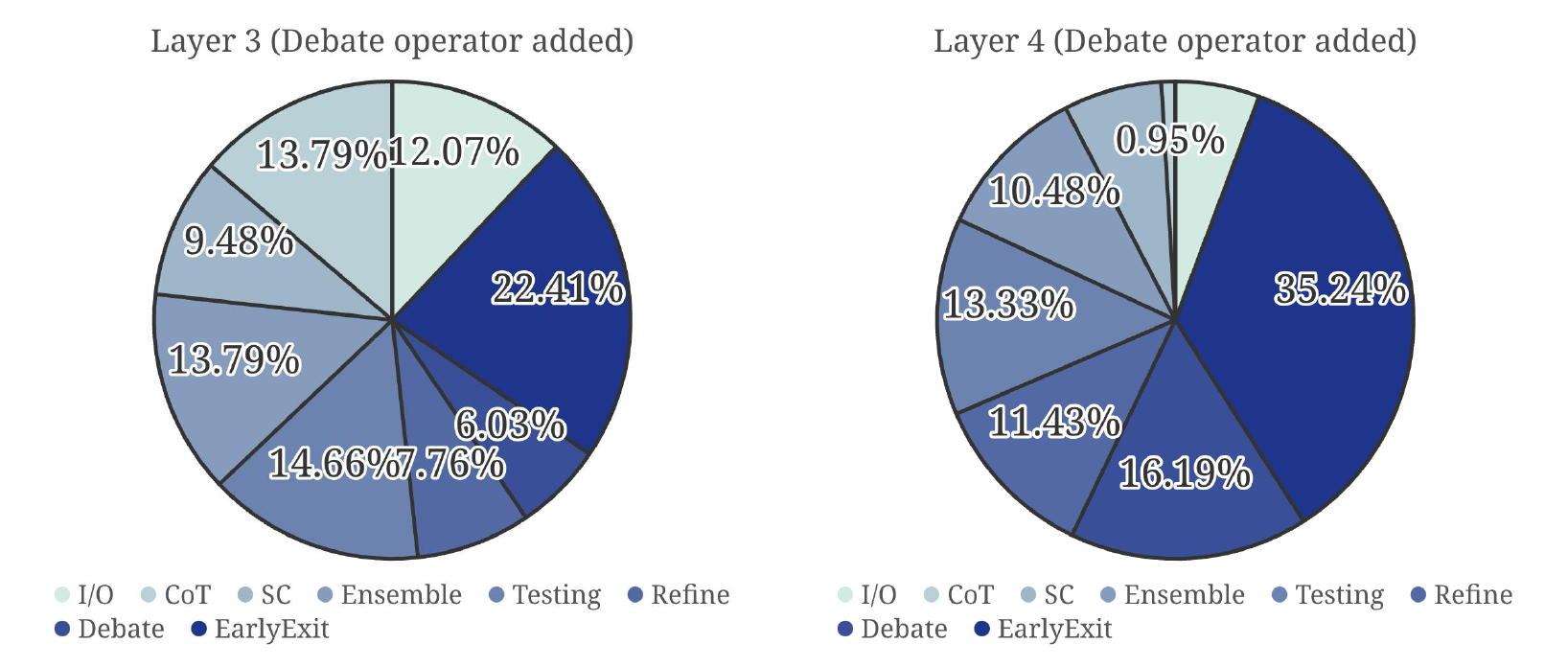}
\vspace{-1em}
\caption{The layer-wise distribution of \ourmethod on HumanEval benchmark with Debate operator. Note that the agentic supernet is optimized with other operators, while the Debate operator is introduced only during the inference stage. It can be observed that, despite not being exposed to this operator during training, \ourmethod can still reasonably select it during the multi-agent architecture sampling process.
}
\vspace{-0.5em}
\label{fig:dynamic2}
\end{figure*}

\end{document}